\documentclass[10pt,journal,compsoc]{IEEEtran}
\usepackage{amsmath,amsfonts}
\usepackage{algorithmic}
\usepackage{array}
\usepackage{textcomp}
\usepackage{stfloats}
\usepackage{url}
\usepackage{verbatim}
\usepackage{graphicx}
\usepackage{multirow}
\usepackage{graphicx}
\usepackage{subfloat}
\usepackage{subfigure}
\usepackage{booktabs}
\usepackage{utfsym}
\usepackage{xcolor,colortbl}
\usepackage{amsmath}
\usepackage{float}
\usepackage{pifont}
\definecolor{gray}{gray}{0.5}

\usepackage[lined,boxed,commentsnumbered,ruled]{algorithm2e}

\usepackage[nocompress]{cite}

\usepackage[colorlinks,
            linkcolor=black,
            anchorcolor=black,
            urlcolor=red,
            citecolor=black]{hyperref}

\def\BibTeX{{\rm B\kern-.05em{\sc i\kern-.025em b}\kern-.08em
    T\kern-.1667em\lower.7ex\hbox{E}\kern-.125emX}}
\usepackage{balance}

\begin{document}

\title{Towards Real-World Aerial Vision Guidance with Categorical 6D Pose Tracker}

\author{Jingtao~Sun,~\IEEEmembership{Member,~IEEE,}
        Yaonan Wang,~\IEEEmembership{Member,~IEEE,}
        Danwei Wang,~\IEEEmembership{Life Fellow,~IEEE}
\IEEEcompsocitemizethanks{\IEEEcompsocthanksitem Jingtao Sun, Yaonan Wang are with the College of Electrical and Information Engineering and the National Engineering Research Centre for Robot Visual Perception and Control, Hunan University, Changsha 410082, China. {E-mail: \{jingtaosun, yaonan\}}@hnu.edu.cn.
\IEEEcompsocthanksitem Danwei Wang is with the School of Electrical and Electronic Engineering,
Nanyang Technological University, SG 639798, Singapore. E-mail: edwwang@ntu.edu.sg.
}
\thanks{(Corresponding author: Yaonan Wang.)}
\thanks{Manuscript received January 9, 2024; revised January 9, 2024.}}

\markboth{Ieee Transactions on Pattern Analysis and Machine Intelligence,~Vol.~14, No.~8, August~2023}%
{Sun \MakeLowercase{\textit{et al.}}: Towards Real-World Aerial Vision Guidance with Categorical 6D Pose Tracker}

\IEEEtitleabstractindextext{%
\begin{abstract}
Tracking the object 6-DoF pose is crucial for various downstream robot tasks and real-world applications. In this paper, we investigate the real-world robot task of aerial vision guidance for aerial robotics manipulation, utilizing category-level 6-DoF pose tracking.  Aerial conditions inevitably introduce special challenges, such as rapid viewpoint changes in pitch and roll and inter-frame differences. To support these challenges in task, we firstly introduce a robust category-level 6-DoF pose tracker (\emph{Robust6DoF}). This tracker leverages shape and temporal prior knowledge to explore optimal inter-frame keypoint pairs, generated under a priori structural adaptive supervision in a coarse-to-fine manner. 
Notably, our \emph{Robust6DoF} employs a Spatial-Temporal Augmentation module to deal with the problems of the inter-frame differences and intra-class shape variations through both temporal dynamic filtering and shape-similarity filtering. We further present a Pose-Aware Discrete Servo strategy (\emph{PAD-Servo}), serving as a decoupling approach to implement the final aerial vision guidance task. It contains two servo action policies to better accommodate the structural properties of aerial robotics manipulation.
Exhaustive experiments on four well-known public benchmarks demonstrate the superiority of our \emph{Robust6DoF}. Real-world tests directly verify that our \emph{Robust6DoF} along with \emph{PAD-Servo} can be readily used in real-world aerial robotic applications.
The project homepage is released at \href{https://github.com/S-JingTao/ICK-Track.git}{Robust6DoF}.
\end{abstract}

}
\maketitle
\IEEEdisplaynontitleabstractindextext
\IEEEpeerreviewmaketitle
\IEEEraisesectionheading{\section{Introduction}\label{sec:introduction}}
\IEEEPARstart{T}{racking} object Six Degree-of-Freedom (6-DoF) pose is one of the most fundamental tasks in computer vision and robotic applications, such as manipulation~\cite{zhu2023toward}, aerial tracking~\cite{huang2023anti,cao2023towards,ma2023adaptive} and navigation. Pioneering works in object 6-DoF pose tracking mostly adopt the standard format, where the 3D CAD model of the object instance is used to achieve remarkable accuracy, often referred as \emph{instance-level 6-DoF pose tracking}. However, acquiring the prefect 3D model is challenging in realistic settings. In this end, we focus on the more demanding study of the problem of \emph{aerial category-level 6-DoF pose tracking}. The objective is to real-time estimate the 6-DoF pose of novel object instances within any one category in the aerial down-look scene, while assuming that 3D geometry model of the instance is unavailable. Furthermore, visual tracking-based methods have drawn considerable attention for unmanned aerial vehicles (UAVs), such as aerial cinematography, visual localization and geographical survey. In this work, we also aim to develop these aerial category-level pose tracking apporaches to tackle the visual guidance task in the field of UAVs, especially for aerial robotics manipulation. The goal of this task is to allow aerial robot to self-guide to the static object or actively follow the moving target.

To date, most currently available category-level 6-DoF pose tracking methods~\cite{wang20206,weng2021captra,sun2022ick,lin2022keypoint,yu2023cattrack} adopt either headmap-based pipline or tracking-by-detection framework, \emph{e.g.,} 6-PACK~\cite{wang20206}. However, these methods neglect the strong correlations inherently existing among consecutive frames, such as inter-frame differences, making it challenging for them to capture changes in the camera's viewpoint over time. Consequently, these pose trackers or estimators do not work as well in aerial scenarios where the captured object image data may exhibit severe perspective drifting. These are caused by different complex conditions, such as high-speed motions and occlusions in an aerial bird's-eye view. In conclusion, aerial category-level 6-DoF pose tracking faces several challenges: 1) The intra-class shape variations within same category, a major challenge that remains limited so far. Canonical/normalizated spaces were introduced in prior works~\cite{wang2019normalized,chen2020learning} to address this issue, and several other methods~\cite{tian2020shape,chen2021sgpa} employed a shape prior to adapt shape inconsistency within the same category, \emph{etc}. However, these methods lack an explicit temporal representation between continuous frames, limiting their performance for aerial pose estimation; 2) The inter-frame differences. Aerial conditions inevitably introduce special challenges including motion blur, camera motion, occlusion and so on. In particular, fast-changing views in pitch and roll hinder the pose tracking performance in aerial scenes. To our knowledge, existing methods do not account for this situation; 3) the limited computing power of aerial platforms restricts the deployment of time-consuming state-of-the-art methods. Hence, an ideal tracker for aerial 6-DoF pose tracking must be both robust and efficient. 

\begin{figure*}
    \centering
    \subfigure[The configuration of guidance task for aerial robotics manipulation.]
    {    \centering
        \includegraphics[width=0.48\textwidth]{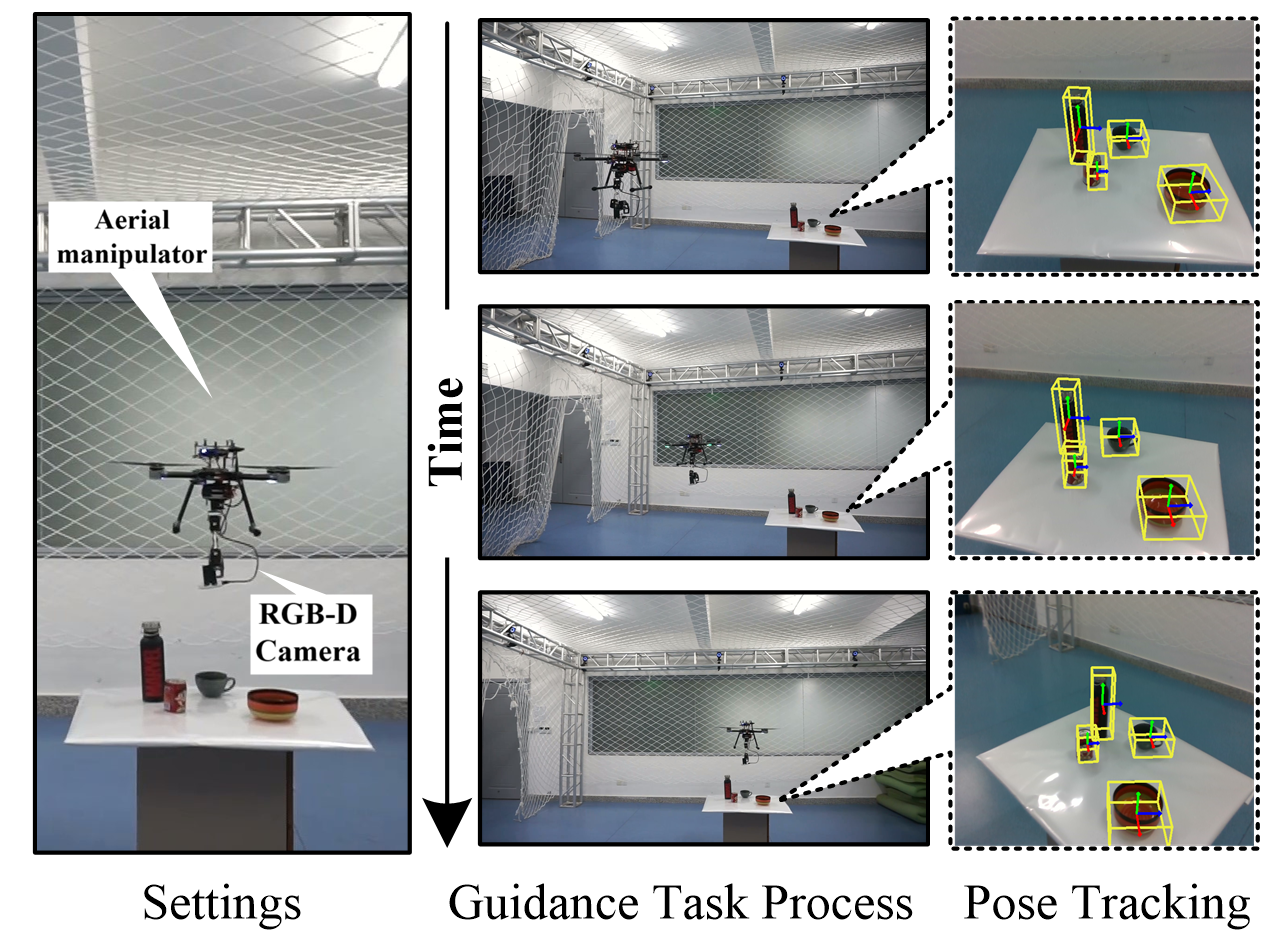}
                
    }
    \subfigure[Comparison of \emph{Robust6DoF} with representative SoTA baselines.]
    {    \centering
        \includegraphics[width=0.45\textwidth]{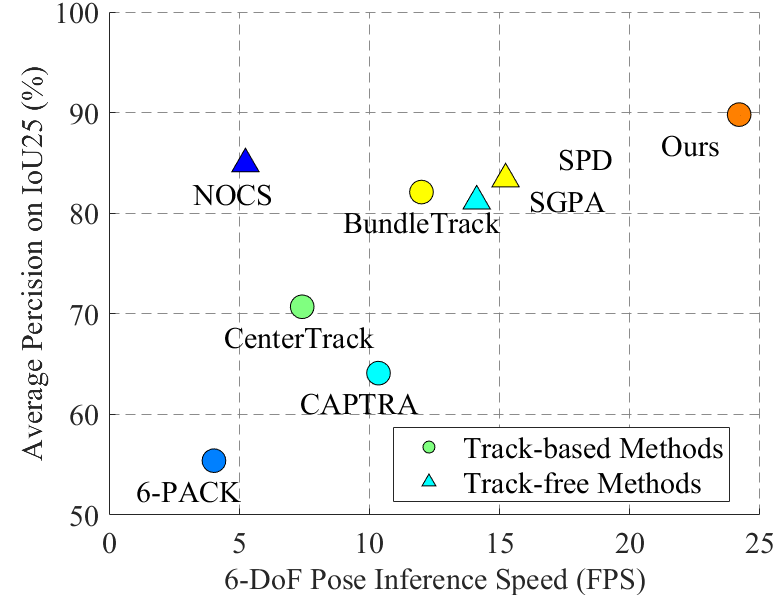}
           
    }
    \caption{\textbf{The overall introduce of pipeline.} \textbf{a)} By following the real-time 6-DoF pose tracking generated from our \emph{Robust6DoF}, the aerial manipulator gradually begins to self-guide to the desired position where the targeted object's pose is infinitely close to the desired value. \textbf{b)} Proposed \emph{Robust6DoF} achieves top performance on the metric of $IoU25$ with the best inference speed on public NOCS-REAL275 dataset. We test competitive category-level track-based and track-free (single pose estimation) methods utilizing their offical checkpoints and codes, respectively. All results are measured on the same device to be fair.}\label{fig_all}\vspace{-0.3cm}
\end{figure*}
As for the following vision guidance task for aerial robotics manipulation, a significant challenge arises from the inherent instability of UAVs. The mounting of an onboard manipulator further increases the nonlinearity of the UAV system. This complexity renders traditional 2D visual-based technologies suboptimal for for solving aerial vision guidance task.
Visual servoing technology is particularly appealing to us due to its scalability in applications and practicality in operations. However, previous related works~\cite{chen2024adaptive,he2023image,hay2023noise} maninly focus on Image-Based Visual Servoing (IBVS) and do not address the problem of the object's 6-DoF pose, where the pose state of the targeted object is usually assumed to be known. Unlike these methods, we propose a robust and efficient pose tracker and utilize its real-time object's 6-DoF pose information to achieve the vision guidance task via a decoupling servo strategy. 

To address the aforementioned challenges, we propose an pose-driven technology to accomplish the aerial vision guidance task for aerial robotics manipulation. As shown in the right of Fig.~\ref{fig_all} (a), to solve primary problem in this task, namely aerial category-level 6-DoF pose tracking, we firstly introduce a robust category-level 6-DoF pose tracker named \textbf{\emph{"Robust6DoF"}}, employing a three-stage pipeline. At the \textbf{first stage}, to conduct the object-aware aggregated descriptor with point-pixel information, our \emph{Robust6DoF} employs a 2D-3D Dense Fusion Transformer to learn dense per-point local correspondences for arbitrary objects in the current observation. At the \textbf{second stage}, unlike related category-level methods in~\cite{wang20206,weng2021captra,sun2022ick,lin2022keypoint,yu2023cattrack}, we present a Shape-Based Spatial-Temporal Augmentation module to address the challenge of inter-frame differences and intra-class shape variations. This module employs both temporal dynamic filtering and shape-similarity filtering with an encoder-decoder structure. In this way, these aggregated descriptor is updated to a group of augmented embeddings. At the final \textbf{third stage}, a Prior-Guided Keypoints Generation and Match module is proposed to seek the optimal inter-frame key-point pairs based on these augmented embeddings. Specifically, we apply a priori structural adaptive supervision mechanism in a coarse-to-fine manner to enhance the robustness of the generated keypoints. The final 6-DoF pose is obtained using the Perspective-n-Point algorithm and RANSAC.

As displayed in the middle of Fig.~\ref{fig_all} (a), to tackle second problem in this task, namely visual guidance for aerial robot, we further propose a Pose-Aware Discrete Servo Policy called \textbf{\emph{"PAD-Servo"}}, including two servo action schemes: \textbf{(1)} Rotational Action Loop, generates the rotational action signal for onboard manipulator in 3D Cartesian space. This signal is derived from the rotation matrix of 6-DoF pose tracked by our \emph{Robust6DoF}; \textbf{(2)} Translation Action Loop, produces the translational action signal for aerial vehicle in 2D image space, This signal comes from the location vector of our \emph{Robust6DoF}'s pose tracking results. 
This separated design can be perfectly adapted to the aerial robot's kinematic model to realize the collaborative actions for both aerial vehicle and onboard manipulator.

We evaluate the performance of our \emph{Robust6DoF} on four publicly available datasets and achieves state-of-the-art results. It's noteworthy that our \emph{Robust6DoF} achieves top performance on the metric of $IoU25$ along with the best tracking speed in NOCS-REAL275 dataset, as depicted in Fig.~\ref{fig_all} (b). Furthermore, we conduct the real experiment in a real-world aerial robotic platform to validate the practicality of our \emph{PAD-Servo} using trained \emph{Robust6DoF} and realize robust real-world results.
The original contributions of this paper can be summarized as follows:

\begin{itemize}
\item To address the problem of aerial category-level 6-DoF pose tracking, we introduce a robust category-level 6-DoF pose tracker utilizing temporal and shape prior knowledge, along with a priori structural adaptive supervision mechanism for keypoint pairs generation. To our best knowledge, we are the first to solve the problem of aerial category-level object 6-DoF pose tracking in aerial high-mobility scenario.
\item To tackle the challenges of inter-frame differences and intra-class variations, we present a Shape-Based Spatial-Temporal Augmentation module through both temporal dynamic filtering and shape-similarity filtering. It improves the robustness of pose tracking for different instances in real-time aerial scene.
\item From the robotic system view, we design an efficient Pose-Aware Discrete Servo strategy to achieve the visual guidance task for aerial robotics manipulation, that is fully adapted to our \emph{Robust6DoF} pose tracker.
\item We conduct a series of experiemnts on NOCS-REAL275~\cite{wang2019normalized}, YCB-Video~\cite{xiang2017posecnn}, YCBInEOAT~\cite{wen2020se} and Wild6D~\cite{ze2022category} datasets. Our \emph{Robust6DoF} achieves new state-of-the-art performance. Moreover, the real-world experiemnts show that the feasibility of our techniques in realistic aerial robotics scenes.

\end{itemize}

The remainder of this article is organized as follows. In the next section, we discuss the related works. Sec.~\ref{sec:statement} analyses the notation and task description. Sec.~\ref{sec:method} describes the proposed approach and its core modules. The experiments are reported in Sec.~\ref{sec:experiment}. Finally, we summarizes the proposed method's limitations, discusses future work and concludes the paper in Sec.~\ref{sec:conclusion}.

\section{Related Work}\label{sec:related}
This section will review the related works on aerial object tracking, 6-DoF pose estimation and tracking, and the visual servoing for aerial robotics manipulation, respectively.
\subsection{Aerial Visual Object Tracking}
Visual object tracking can be boardly divided into three categories: Siamese-based, DCF-based and Transformer-based. In recent, several DCF-based methods have been deployed for aerial visual object tracking, including SARCT~\cite{xue2020semantic}, ARCF~\cite{huang2019learning}, MRCF~\cite{ye2021multi}. In~\cite{xue2020semantic}, Xue \emph{et al.} presented a semantic-aware correlation approach with low computing cost to enhance the performance of DCF-tracker. Another representative efforts include AutoTrack~\cite{li2020autotrack} and TCTrack~\cite{cao2022tctrack}. Most of these methods continuously update the model  from past historical information. In these aerial tracking methods, the targets are relatively small and are often in a state of fast motion. In this work, we aim to track the tabletop objects that have a big size in camera's view. Additionally, aerial 6-DoF pose tracking is more challenging due to the rapid viewpoint changes in pitch and roll.
\subsection{Object 6-DoF Pose Estimation}
Early advancements in 6-DoF pose estimation can be broadly categorized into two groups: \emph{instance-level}~\cite{cao2023dgecn++,jiang2023center,zhao2023learning,zhou20233d,chen2023activezero++,wang2021occlusion,wang2021geopose,shugurov2021dpodv2} and \emph{category-level}~\cite{zou2023gpt,zou20226d,chen2021sgpa,wang2019normalized,lin2023vi,lin2021dualposenet,chen2021fs,chen2020learning,lin2023vi,liu2022toward,lin2022sar,wang2023query6dof,tian2020shape,yu2023category,lee2022uda,lee2023tta,zheng2023hs}. Instance-level methods predict object pose using known 3D CAD models and can also be classified into template-based methods and feature-based methods. However, obtaining accurate CAD models of unseen objects is a challenge for these type of methods. 
In contrast, category-level methods aim to predict the pose of instances without specific models. NOCS~\cite{wang2019normalized} pioneered direct regression of canonical coordinates for each instances, while CASS~\cite{chen2020learning} developed a variational autoencoder for reconstructing object models. 
Category-level methods still face challenges due to RGB or RGB-D features sensitivity to surface texture and the problem of the intra-shape variation. 

\subsection{Object 6-DoF Pose Tracking}
Object 6-DoF pose tracking is an important task in robotics and computer vision. Researchers have primarily concentrated on this task in two distinct manners: \emph{(1) instance-level pose tracking}:
This type of approach relies on the complete instance's 3D CAD models, and the notable efforts include PA-Pose~\cite{liu2023pa}, Deep-AC~\cite{wang2023deep}, BundleSDF~\cite{wen2023bundlesdf}, PoseRBPF~\cite{9363455} and so on~\cite{wen2020se}; \emph{(2) category-level pose tracking}: This type of method operates without specific 3D model requirements. 
Wang~\emph{et al.}~\cite{wang20206} first introduced a novel category-level tracking benchmark,  constructing a set of 3D unsupervised keypoints for pose tracking, named 6-PACK. To address per-part pose tracking for articulated objects, CAPTRA~\cite{weng2021captra} presented an end-to-end differentiable pipline without any pre- or post- processing.
ICK-Track~\cite{sun2022ick} also introduced a inter-frame consistent keypoints generation network to generate the corresponding keypoint pairs in pose tracking.
Lin~\emph{et al.} proposed CenterTrack~\cite{lin2022keypoint}, using the CenterNet framework to achieve the categorical pose tracking. In~\cite{yu2023cattrack}, Yu~\emph{et al.} proposed CatTrack to solve this problem with the single-stage keypoints-based registration.

Our proposed \emph{Robust6DoF} is also a category-level 6-DoF pose tracking method. However, different from existing category-level track-based methods~\cite{wang20206,weng2021captra,sun2022ick,lin2022keypoint,yu2023cattrack}, we address the challenges of the intra-shape and inter-frame variations by leveraging both temporal prior and the shape prior knowledge. Meanwhile, we also consider the inter-frame key-points generation under the supervision of canonical shape priors, facilitating real-time adaptation of generated key-points to variations in inter-frame differences and distinguishes between observation and shape prior. Notably, Our method stands as the pioneering solution to the special aerial challenge in category-level 6-DoF pose tracking.
\vspace{-0.4cm}
\subsection{Visual Servoing for Aerial Robotics Manipulation}
The standard solution to the visual servoing task relies on Position-Based Visual Servo (PBVS) or Image-Based Visual Servoing (IBVS). IBVS is more robust than PBVS in handling uncertainties and disturbances that affect the robot's model, has proven to be a viable method for addressing aerial robotics manipulation tasks~\cite{chen2024adaptive,he2023image,santamaria2017uncalibrated,kim2016vision,hay2023noise}.
In~\cite{chen2024adaptive}, Chen \emph{et al.} introduced an robust adaptive visual servoing method to achieve a compliant physical interaction task for aerial robotics manipulation. 
In~\cite{hay2023noise}, Oussama \emph{et al.} proposed to use a deep neural networks (DNNs) for visual servoing applications of UAVs.
In~\cite{kim2016vision}, a typical vision guidance system based on IBVS was integrated with passivity-based adaptive control for aerial robotics manipulation, showcasing promising results in simulation experiments and indicating potential real-world applications. 
Additionally, other methods have been developed to address visual-based tasks in aerial robotics manipulation, such as~\cite{gabellieri2020compliance,zhang2019robust}. 
Although IBVS-based techniques are well-established, they exhibit insensitivity to manipulator calibration and susceptibility to local optima. 
In our work, we advance the field by leveraging both PBVS and IBVS methods and directly generates the movement actions of the aerial vehicle and manipulator through their respective servo action loops. Our proposed method is better adapted to the nonlinear nature of aerial manipulator.

\newpage

\begin{figure}[t]

    \centering
    \includegraphics[width=0.9\columnwidth]{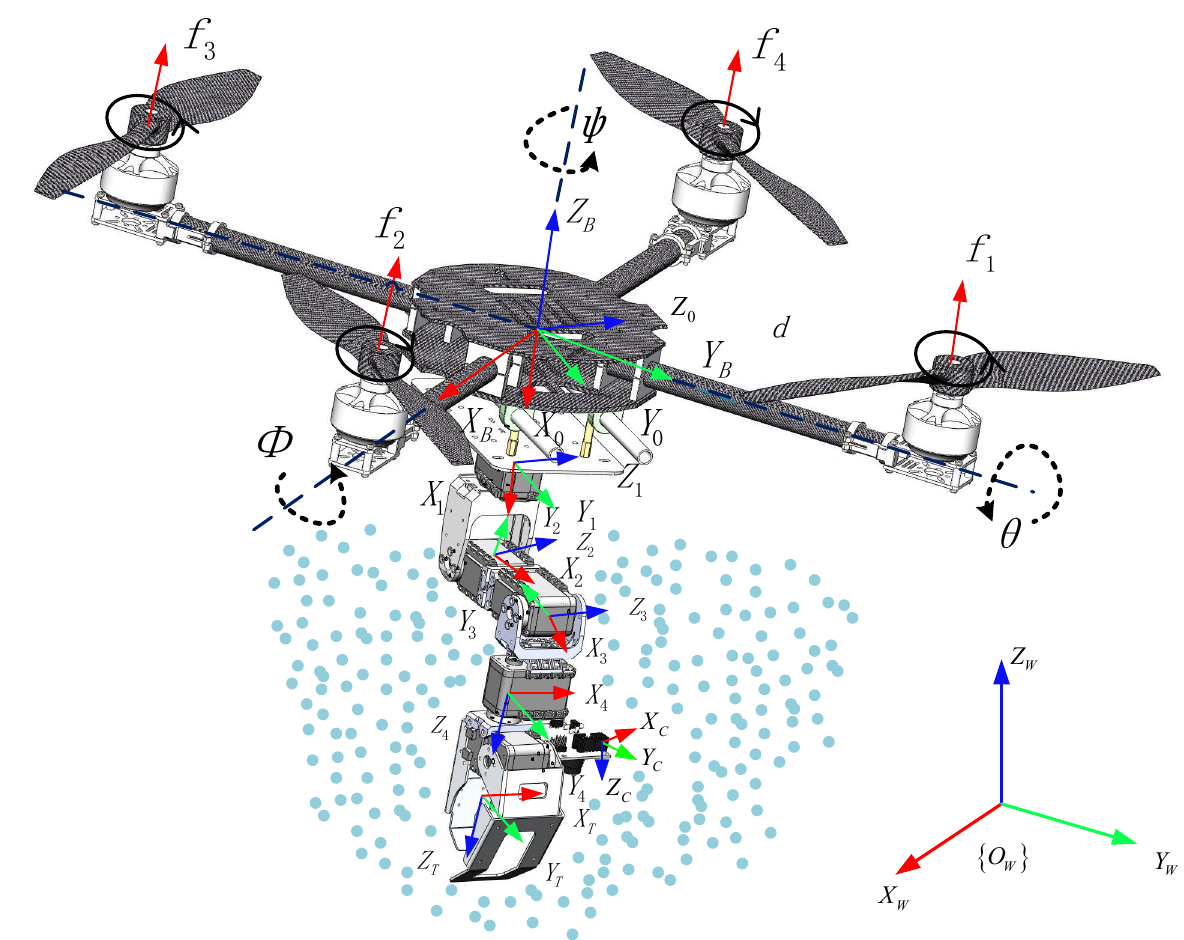}
    \caption{\textbf{Establishment of aerial robot frame.} ${\rm{\{ W : }}{{\rm{O}}_W}{\rm{ - }}{{\rm{X}}_W}{Y_W}{Z_W}{\rm{\} }}$ means the world coordinate frame. ${\rm{\{ B : }}{{\rm{O}}_B}{\rm{ - }}{{\rm{X}}_B}{Y_B}{Z_B}{\rm{\} }}$ means the base coordinate frame of the aerial vehicle. ${\rm{\{ L_i : }}{{\rm{O}}_i}{\rm{ - }}{{\rm{X}}_i}{Y_i}{Z_i}{\rm{\} }}$ means the body frame of the $i$ link of robotic manipulator (i = 0,1,2,3,4), where $i = 0$ indicates the base frame of manipulator. ${\rm{\{ T : }}{{\rm{O}}_T}{\rm{ - }}{{\rm{X}}_T}{Y_T}{Z_T}{\rm{\} }}$ means the cooridate frame of the actuator. ${\rm{\{ C : }}{{\rm{O}}_C}{\rm{ - }}{{\rm{X}}_C}{Y_C}{Z_C}{\rm{\} }}$ means the onboard camera frame. The blue dot represents the 3D work space of onboard manipulator.}\label{FIG_1}
    
\end{figure}

\section{Preliminary and task statement}\label{sec:statement}
\subsection{Robot Frame and Velocity Transmission}
In our work, we use a general aerial manipulator as the robotics platform, consisting of a multirotor UAV with a 4-link serial robotic manipulator and an RGB-D camera configured as eye in hand.
The schematic diagram and the reference frames of this platform are shown in Fig. \ref{FIG_1}.
We define the following representations: $\dot p = {({v_x},{v_y},{v_z})} \in {\mathbb{R}^3}$ and $\omega  = {({\omega _x},{\omega _y},{\omega _z})} \in {\mathbb{R}^3}$, representing the linear and angular velocities of the aerial vehicle respectively. And $\dot \eta  = {({\dot \eta _1},{\dot \eta _2},{\dot \eta _3},{\dot \eta _4})} \in {\mathbb{R}^4}$ denotes the joint angular velocity of the onboard manipulator. 
An expression that eventually contains all generalized velocities is given by:
\begin{align}
q = {({\dot p^T},{\omega ^T},{\dot \eta ^T})}.
\end{align}

Following the differential kinematic propagation, the velocity transmission between all generalized velocities and the velocity of the onboard camera can be derived as: 

\begin{align}\label{trans}
{V_C} = Jq^T,
\end{align}
where ${V_C} = {{[^c}{{\dot p}^T}_c{,^c}\omega _c^T]^T} \in {\mathbb{R}^{6 \times 1}}$ is the velocity vector of camera frame expressed in its own frame, consisting of linear and angular velocities. The generalized Jacobian matrix $J \in {\mathbb{R}^{6 \times 10}}$ is given by:
\begin{align}
\resizebox{.9\hsize}{!}{$
J = \left[ {\begin{array}{*{20}{c}}
{{{(U_C^B)}^T}}&{{{(U_C^{{L_1}})}^T}z}&{{{(U_C^{{L_2}})}^T}z}&{{{(U_C^{{L_3}})}^T}z}&{{{(U_C^{{L_4}})}^T}z}
\end{array}} \right],$}
\end{align}
where $z = {\left[ {\begin{array}{*{20}{c}}
{{0_{1 \times 5}}}&1
\end{array}} \right]^T}$. 
And $U_\alpha ^\beta$ is the generalized transformaton matrix between any two adjacent cooridnate frames $\{ \alpha \} $ and $\{ \beta \} $, expressed as:
\begin{align}
U_\alpha ^\beta  = \left[ {\begin{array}{*{20}{c}}
{R_\alpha ^\beta }&{{0_{3 \times 3}}}\\
{P(r_{\beta ,\alpha }^\beta )R_\alpha ^\beta }&{R_\alpha ^\beta }
\end{array}} \right],
\end{align}
where ${R_\alpha ^\beta }$ is the rotation matrix between frame ${ \left\{  \alpha  \right\} }$ and frame $\left\{  \beta  \right\}$, and $r_{\beta ,\alpha }^\beta  = ({r_x},{r_y},{r_z})$ is the translation vector of frame $\{ \alpha \} $ with respect to and expressed in frame $\{ \beta \} $. Here, $\{ \alpha ,\beta \} $ belongs to the any pair of all body-fixed frames in the system, \emph{i.e.,} $\{ \alpha ,\beta \}  \in \{ B,{L_1}, \ldots ,C\}$, as depicted in Fig. \ref{FIG_1}. 
After calculation, the Eq.~(\ref{trans}) can also be expressed as:
\begin{align}
\resizebox{.9\hsize}{!}{$
\left[ {\begin{array}{*{20}{c}}
{{}^c{{\dot p}_c}^T}\\
{{}^c{\omega _c}^T}
\end{array}} \right] = \left[ {\begin{array}{*{20}{c}}
{{{(R_C^B)}^T}}&{{0_{3 \times 3}}}&{{0_{3 \times 4}}}\\
{{{(P(r_{B,C}^B)R_C^B)}^T}}&{{{(R_C^B)}^T}}&{{T_{3 \times 4}}}
\end{array}} \right]\left[ {\begin{array}{*{20}{c}}
{{{\dot p}^T}}\\
{{\omega ^T}}\\
{{{\dot \eta }^T}}
\end{array}} \right],$}
\end{align}
where ${{T_{3 \times 4}}}$ is a matrix composed of the last row of ${R_\alpha ^\beta }$.
\begin{figure*}[ht]
    \centering
    \includegraphics[width=\textwidth]{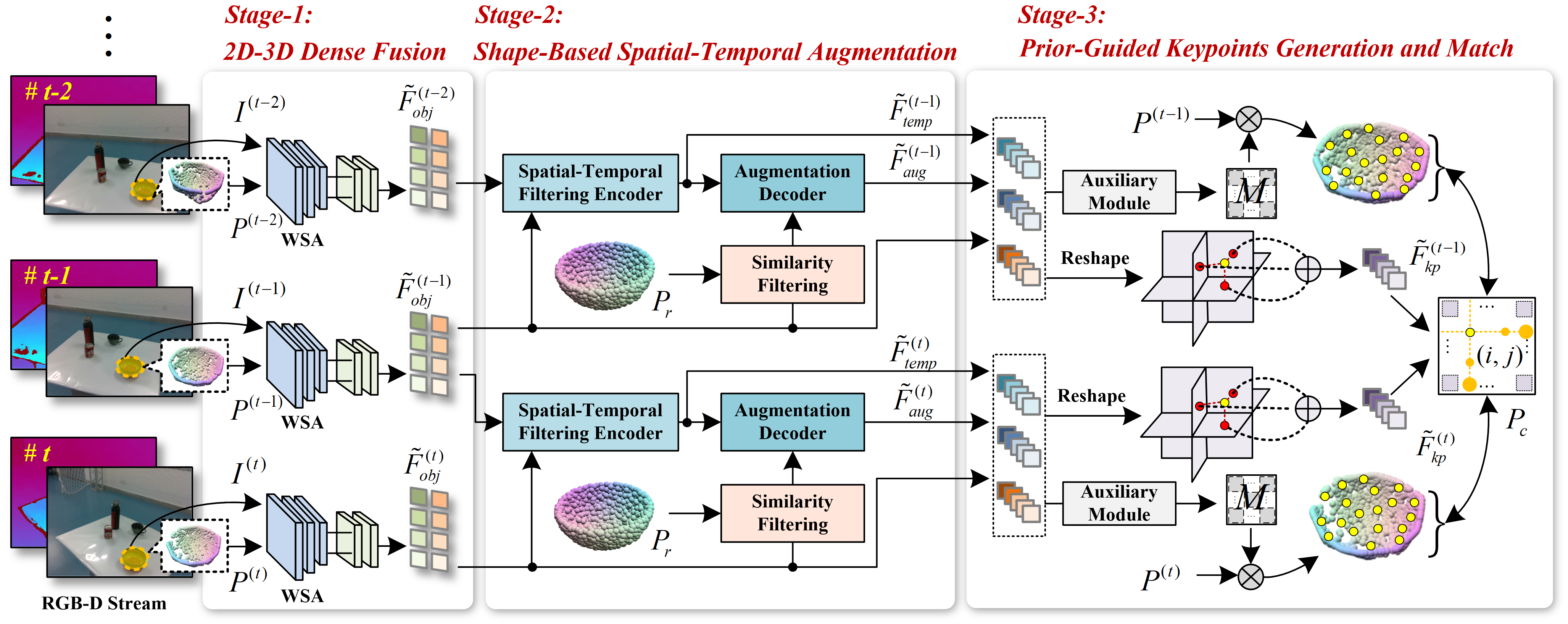}\vspace{-0.2cm}
    \caption{\textbf{Complete framework of our category-level 6-DoF pose tracker termed~\emph{Robust6DoF}}. It takes RGB-D video stream captured by the onboard camera as input, and tracks the 6-DoF pose ${{\cal P}^{(t)}}$ of the arbitrary object in the current observation. It mainly consists of three phases. \textbf{Stage-1}: 2D-3D dense fusion for pixel-point object's local descriptor $\tilde F_{obj}^{(t)}$ aggregation (shown in Fig.~\ref{FIG_structure}~(a)); \textbf{Stage-2}: shape-based spatial-temporal augmentation is employed for comprehensive refinement to obtain a group of embeddings $\{ \tilde F_{obj}^{(t)},\tilde F_{temp}^{(t)},\tilde F_{aug}^{(t)}\} $, taking advantage of both temporal prior and shape prior knowledge (shown in Fig.~\ref{FIG_structure}~(b) and (c)); and \textbf{Stage-3}: prior-guided keypoints generation and matching for $n$ inter-frame keypoints $(k_i^{(t - {\rm{1}})},k_j^{(t)})$ construction and accurate alignment in a coarse-to-fine manner. Utilizing these optimally matched keypoint pairs, we solve for the final object’s 6D pose using the PnP and RANSAC algorithms.}\label{FIG_overview}
    \vspace{-0.4cm}
\end{figure*}
\subsection{Task Description}
Eq. (5) clearly shows that the linear velocity of the onboard camera is a result of the linear velocity of the aerial vehicle. Similarly, considering the underactuation of the aerial vehicle in two degrees of freedom (${ \omega _x}$ and ${ \omega _y}$), the angular velocity of the camera primarily arises from the motion of each joint of the manipulator (${{{\dot \eta }^T}}$) and ${ \omega _z}$. In other word, hidden relationships exist between the linear velocity of the aerial vehicle and the camera's translation ($T$), as well as between the angular velocity of the manipulator and the camera's rotation ($R$). Thus, the mentioned visual guidance task can be decomposed into two processes, namely, \textbf{(1)} \emph{6-DoF pose tracking for object}, and \textbf{(2)} \emph{visual servoing for aerial robot}.
Based on the above analysis and the formulation expressed in the right of Fig.\ref{fig_all}, we can address this task as follows: our method first take the RGB-D video stream captured by the onboard camera as inputs to real-time tracking object's 6-DoF pose and subsequently generate the servo action signals for the aerial vehicle and onboard manipulator, respectively:

\begin{equation}
\resizebox{0.7\hsize}{!}{$
\begin{aligned}
\left\{ {\begin{array}{l}
{{{\cal P}^{(t)}} = [{R^{(t)}}|{T^{(t)}}] = \Delta {{\cal P}^{(t)}} \cdot  \ldots  \cdot {{\cal P}^{(0)}}}\\
{{{\dot \eta }^{(t)}} = {\partial_{rot}} \mapsto \mathop {\min }\limits_{{\varepsilon _r}} [{R^{(t)}},{R^*}]}\\
{{v^{(t)}} = {\partial_{tra}} \mapsto \mathop {\min }\limits_{{\varepsilon _p}} [{T^{(t)}},{T^*}]}
\end{array}} \right.,
\end{aligned}$}
\end{equation}
where ${\partial_{rot}}$ and ${\partial_{tra}}$ are denotes the servo action function. The change of pose $\Delta {{\cal P}^{(t)}} \in \bf{SE}(3)$ contains the change in rotation $\Delta R^{(t)} \in \bf{SO}(3)$ and the change in translation $\Delta T^{(t)} \in {\mathbb{R}^3}$, $\Delta {{\cal P}^{(t)}} = [\Delta {R^{(t)}}|\Delta {T^{(t)}}]$. The absolute 6-DoF pose ${{\cal P}^{(t)}} = [{R}^{(t)}|{T}^{(t)}]$ in current observation can then be derived by recursing the previous tracking results over time. The initial pose ${{\cal P}^{(0)}}$ is set to the estimated pose state at the beginning of the guidance task.

\section{Approach}\label{sec:method}
In this section, we will first introduce our proposed category-level object 6-DoF pose tracker, namely \textbf{\emph{Robust6DoF}} in Sec.~\ref{sec:method_a} and then present the detailed scheme of our Pose-Aware Discrete Servo Policy called \textbf{\emph{PAD-Servo}} for aerial robotics manipulator in Sec.~\ref{sec:method_b}.


\subsection{Categorical 6-DoF Pose Tracker: Robust6DoF}\label{sec:method_a}

\subsubsection{Network Overview}
In this subsection, we will present an overview of our proposed category-level 6-DoF pose tracker and then provide detailed introductions to each component in our designed network. As depicted in Fig.~\ref{FIG_overview}, our goal is to estimate continuously the change of the 6-DoF pose, denoted as $\Delta {{\cal P}^{(t)}}$, for the target object within an arbitrary known category. The core inputs of our network include the observed RGB-D video stream captured by the onboard camera and the corresponding categorical shape prior ${P_r} \in {\mathbb{R}^{{N_r} \times 3}}$, which is converted in advance into the same coordinate as the camera. 
For simplicity, the number of shape prior models is uniformly sampled to be consistent with ${P^{(t)}}$,~\emph{w.r.t.,} ${N_r} = N$.
Different from the recent methods~\cite{wang20206,weng2021captra,sun2022ick,lin2022keypoint,yu2023cattrack} for category-level pose tracking, we employ a three-stage pipline, as displayed in Fig.~\ref{FIG_overview}. \textbf{\emph{stage-1:}} We first integrate the local pixel-point dense feature descriptor for each target object using the proposed 2D-3D Dense Fusion Transformer (Sec.~\ref{s-1}); \textbf{\emph{stage-2:}} Subsequently, we introduce a Shape-Based Spatial-Temporal Augmentation module with a encoder-decoder structure to dynamically enhance this object-aware descriptor utilizing both temporal prior and shape prior knowledges. It ensures the adaptability of final augmented representations to inter-class variations and inter-frame differences (Sec.~\ref{s-2}); \textbf{\emph{stage-3:}} All enhanced embeddings are passed through the proposed Prior-Guided Keypoints Generation and Match module to build the 3D-3D inter-frame keypoint pair correspondences in a coarse-to-fine manner (Sec.~\ref{s-3}). The final pose tracking is solved with the Perspective-n-Point (PnP) algorithm and RANSAC using these generated and aligned sets of keypoint pairs.

\subsubsection{2D-3D Dense Fusion Transformer}\label{s-1}
\begin{figure*}[ht]
    \centering
    \includegraphics[width=0.9\textwidth]{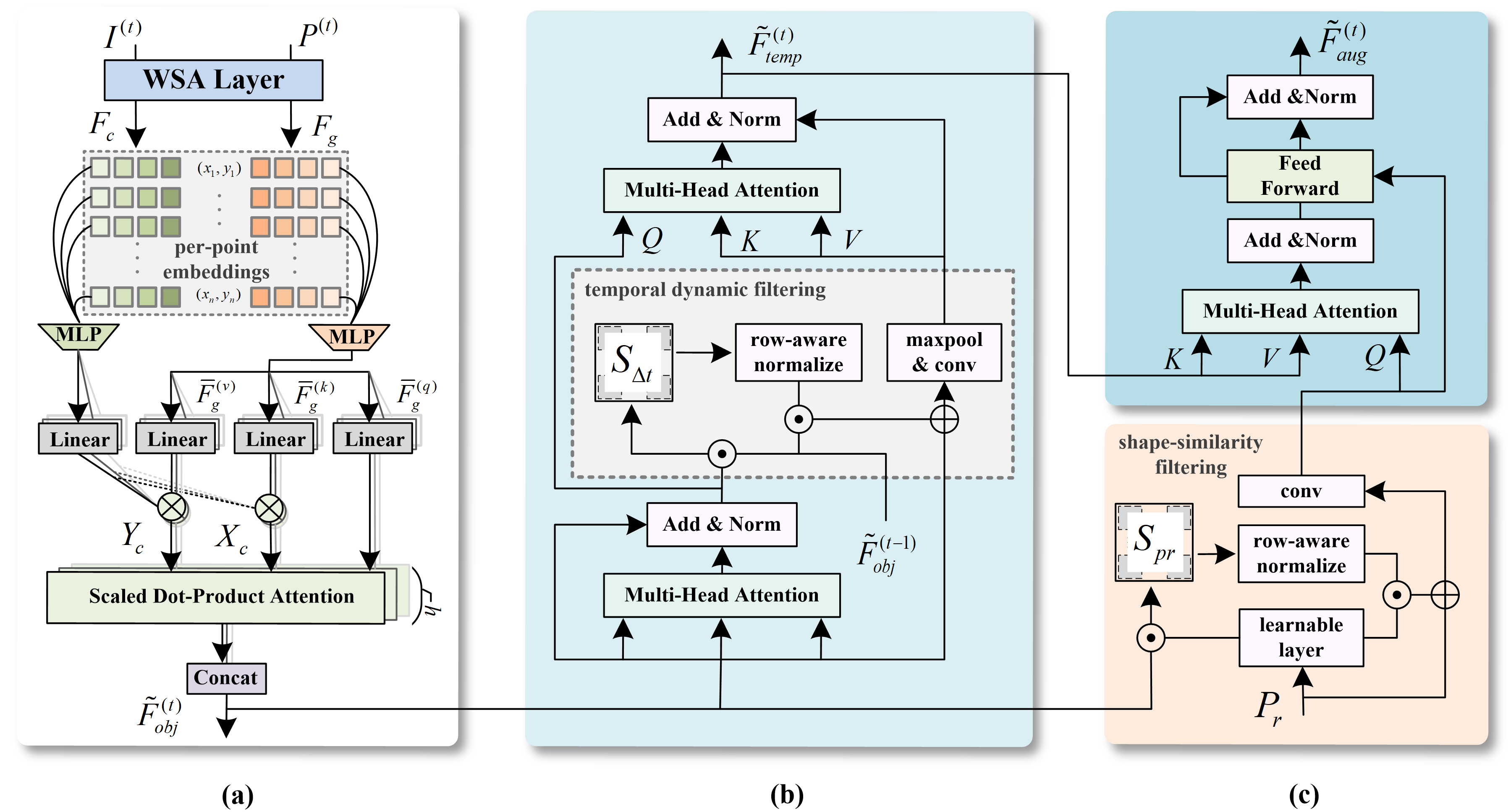}\vspace{-0.2cm}
    \caption{\textbf{Detailed structure of the tracking workflow at the initial two stages.} \textbf{a)}~ 2D-3D Dense Fusion Tramsformer. The image crop and point patch serve as inputs to generate the fused local descriptor $\tilde F_{obj}^{(t)}$ for arbitrary instances in current $t$-th frame. This component primarily consists of two parts: i) The WSA layer is employed for pixel-point dense fusion; ii) The scaled dot-product attention for local feature aggregation. \textbf{b)}~Spatial-Temporal Filtering Encoder. It exploits the temporal knowledge from previous $t-1$-th frame to current one via the proposed temporal dynamic filtering. \textbf{c)}~Augmentation Decoder along with shape-similarity filtering. These blocks leverage the proposed shape-similarity filtering to augment the temporal embedding $\tilde F_{temp}^{(t)}$, effectively addressing the challenge of the intra-category variability.}\label{FIG_structure}
    \vspace{-0.4cm}
\end{figure*}

The objective of this module is to build a local aggregated descriptor for each object by establishing dense per-point feature correspondences between the 3D point patch and the 2D image crop, that serves as the base embeddings for next embedding argumantation.
In earlier works such as~\cite{wang2019densefusion} and~\cite{wang20206}, 2D image and 3D depth information were used separately as inputs without considering the combination of modal-wise features, that resulted in the loss of intermodal correlation during the feature extraction process. In this regard, we present a pixel-point dense fusion module that utilizes the similarity properties of Transformer to enhance the selection of highly correlated feature pairs, as shown in Fig.~\ref{FIG_structure} (a). Concretely, given the current image pixel crop ${I^{(t)}} \in {\mathbb{R}^{H \times W \times 3}}$, along with the observable geometric point patch ${P^{(t)}} \in {\mathbb{R}^{N \times 3}}$ with a one-to-one correspondence through back-projection, we first employ our proposed Weight-Shared Attention (WSA) to map each pixel in the image crop to a color feature embedding ${F_c} \in {\mathbb{R}^{N \times {d_{rgb}}}}$, meanwhile, process the corresponding point in the 3D point patch to a geometric feature embedding ${F_g} \in {\mathbb{R}^{N \times {d_{geo}}}}$. The WSA layer adopts an offset-attention structure:
\begin{align}\label{wsa-1}
\resizebox{.9\hsize}{!}{$
{F_c} = \varphi (\alpha ({\mathcal{F}_q}({I^{(t)}}) \cdot ({\mathcal{F}_k}({P^{(t)}}))^T) \cdot {\mathcal{F}_v}({I^{(t)}}) - {\mathcal{F}_q}({I^{(t)}}))$}
\end{align}
\begin{align}\label{wsa-2}
\resizebox{.9\hsize}{!}{$
{F_g} = \varphi (\alpha({\mathcal{F}_q}({P^{(t)}}) \cdot ({\mathcal{F}_k}({I^{(t)}}))^T) \cdot {\mathcal{F}_v}({P^{(t)}}) - {\mathcal{F}_q}({P^{(t)}}))$}
\end{align}
where $\varphi$ represents the linear and ReLU layer applied to the output features and $\alpha $ denotes the softmax function. ${\mathcal{F}_i},i = q,k,v$ represents the convolutional operation for query, key and value, respectively.

After the initial dense fusion, we aggregate these base dense information and then encode the context-dependent local feature descriptor $\tilde F_{obj}^{(t)}$ for each object in current frame. Inspired by the standpoint proposed by Wang \emph{et al.}~\cite{wang2020linformer}, that the low-rank nature of the context mapping matrix in the self-attention mechanism, we utilize this property not only to reduce the complexity time from $O({N^2})$ to $O(N)$ but also to enhance the instance's pose representation in term of local per-point fusion. Specifically, ${F_c}$ and ${F_g}$ undergo an MLP operation and the color embedding ${F_c}$ is projected into two identical projection matrices $X_{c}, Y_{c} \in {\mathbb{R}^{N \times k}}$. As shown in Fig.~\ref{FIG_structure} (a), we then incorporate them when computing the key and value vectors. This allows us to calculate an ${(N \times k)}$-dimensional context mapping matrix using scaled dot-product attention with multi-heads:
\begin{equation}\label{msa-1}
\resizebox{.9\hsize}{!}{$
\begin{split}
f_g^i &= Attention(\bar F_g^{(q)},\bar F_g^{(k)},\bar F_g^{(v)}) \\
&= Softmax\left( {\frac{{\mathcal{F}(\bar F_g^{(q)}){{({X_c} \cdot \mathcal{F}(\bar F_g^{(k)}))}^T}}}{{\sqrt d }}} \right) \cdot {Y_c} \cdot \mathcal{F}(\bar F_g^{(v)})
\end{split},$}
\end{equation}
\begin{align}
\tilde F_{obj}^{(t)} = Cat(f_g^1,f_g^2, \ldots ,f_g^h),
\end{align}
where ${d}$ is the embedding dimension and $h$ is the number of heads. $\mathcal{F}$ denotes the the linear layer.

\subsubsection{Shape-Based Spatial-Temporal Augmentation}\label{s-2}
Unlike common indoor tabletop scene, the fast change of the onboard camera's view in the aerial brid's-eye perspective, such as pitch or roll, may induce the motion blur or significant inter-frame variations in space scale, \emph{etc}. These challenges are unavoidable in real-time aerial 6-DoF pose tracking. Additionally, the intra-class shape variation with the same class can notably impact the performance of pose tracking for different instances. To our best knowledge, existing category-level pose tracking methods~\cite{wang20206,weng2021captra,sun2022ick,lin2022keypoint,yu2023cattrack} have not completely solved these problems. In this end, we introduce a shape-based spatial-temporal augmentation strategy in this module. This strategy has an encoder-decoder structure leveraging both temporal knowledge dynamic filtering and the shape-similarity filtering, as depicted in the middle of Fig.~\ref{FIG_overview}.

The spatial-temporal filtering encoder aims to solve the challenge of inter-frame differences and initially construct a base temporal embedding $\tilde F_{temp}^{(t)}$ by transforming the prior knowledge from previous frame to current frame. As presented in the Fig.~\ref{FIG_structure} (b), given the local descriptor $\tilde F_{obj}^{(t)} = \{ f_i^{(t)} \in {\mathbb{R}^d}\} _{i = {\rm{1}}}^N$, we first apply a multi-head attention layer to generate ${{\dot F}^{(t)}} = \{ \dot f_i^{(t)} \in {\mathbb{R}^d}\} _{i = {\rm{1}}}^N$:
\begin{align}
{{\dot F}^{(t)}} = Norm(\tilde F_{obj}^{(t)} + MultiHead(\tilde F_{obj}^{(t)},\tilde F_{obj}^{(t)},\tilde F_{obj}^{(t)})),
\end{align}
where the "${Norm}$" indicates the normalization layer. Considering that a faster change in viewpoint results in fewer overlaps among continous frames, making it difficult to capture useful inter-frame information, we need to retain the observable shape features while filtering out irrelevant data.
Therefore, based on $\tilde F_{obj}^{(t - {\rm{1}})} = \{ f_j^{(t - {\rm{1}})} \in {\mathbb{R}^d}\} _{j = {\rm{1}}}^N$ in the previous ${t-1}$-th frame, we compute a spatial-temporal similarity map matrix ${S_{\Delta t}}$ using the vector inner product:
\begin{align}
{S_{\Delta t}}(i,j) =  < \dot f_i^{(t)},f_j^{(t - {\rm{1}})} > ~ \in {\mathbb{R}^{N \times N}},
\end{align}
and then is row-aware normalized through a softmax function to constrained the column elements of ${S_{\Delta t}}$ into the range of $[0,{\rm{1}}]$,~\emph{i.e.,} ${{\bar S}_{\Delta t}} = Softmax{({S_{\Delta t}}(i, \cdot ))_j}$. Where $<, >$ is the inner product. Afterward, we employ a max-pooling operator along with a convolution layer to obtain the filtered descriptor denoted by ${{\ddot F}^{(t)}} = \{ \ddot f_i^{(t)} \in {\mathbb{R}^d}\} _{i = {\rm{1}}}^N$:
\begin{align}
\ddot f_i^{(t)} = MaxPool{(\mathcal{F}([{{\bar S}_{\Delta t}}(i,j) \odot f_j^{(t - {\rm{1}})};f_i^{(t)}]))_j},
\end{align}
where $\mathcal{F}$ denotes the convolution layer and $[;]$ indicates vector concatenation. In this process, we effectively assign weights according to the impact of the previous frame features using the map matrix ${{\bar S}_{\Delta t}}$ and prioritize its the most relevant shape points. With this, the current $t$-th temporal embedding can be obtained as follows:
\begin{align}
\tilde F_{temp}^{(t)} = Norm({{\ddot F}^{(t)}} + MultiHead({{\dot F}^{(t)}},{{\ddot F}^{(t)}},{{\ddot F}^{(t)}})).
\end{align}

To address another challenge of intra-category shape variability, we then employ the canonical shape prior information to augment obtained temporal embedding in the following augmentation decoder. In this end, a shape-similarity filtering block, as depicted in the bottom of Fig.~\ref{FIG_structure} (c), is adopted before the decoding process to adaptively enhance the local object descriptor based on the shape prior ${P_r}$. To jointly optimize the static prior model with the primary network, ${P_r}$
is first fed into a learnable layer to generate the shape-point representation ${F_{pr}} = \{ f_j^{pr} \in {\mathbb{R}^d}\} _{j = {\rm{1}}}^{{N_r}}$. Likewise, a shape-similarity map matrix ${S_{pr}}$ is computed as follows:
\begin{align}
{S_{pr}}(i,j) =  < f_i^{(t)},f_j^{pr} > ~ \in {\mathbb{R}^{N \times {N_r}}},
\end{align}
and then normalized as ${{\bar S}_{pr}}$ using the same operation as before. With this, the filtered features can be denoted by ${{\dddot F}^{(t)}} = \{ \dddot f_i^{(t)} \in {\mathbb{R}^d}\} _{i = {\text{1}}}^N$:
\begin{align}
\dddot f_i^{(t)} = \mathcal{F}([{{\bar S}_{pr}}(i,j) \odot f_j^{pr};{\rho _j}]),
\end{align}
where ${\rho _j}$ is the coordinate of the shape-point. We assign weights according to the impact of shape-point features using the map matrix ${{\bar S}_{pr}}$ to compensate the missing information in current observation. The final augmented output, $\tilde F_{aug}^{(t)}$, is updated by adopting one multi-head attention layer with feed-forward, expressed as:
\begin{equation}
\resizebox{.89\hsize}{!}{$
\begin{split}
&F_{aug}^{(t)} = Norm({{\dddot F}^{(t)}} + MultiHead({{\dddot F}^{(t)}},\tilde F_{temp}^{(t)},\tilde F_{temp}^{(t)}))\\
&\tilde F_{aug}^{(t)} = Norm(F_{aug}^{(t)} + FFN(F_{aug}^{(t)})).
\end{split}$}
\end{equation}
\subsubsection{Prior-Guided Keypoints Generation and Match}\label{s-3}
According to these group of augmented representations $\{ \tilde F_{obj}^{(t)},\tilde F_{temp}^{(t)},\tilde F_{aug}^{(t)}\} $, we now employ these representations to generate the 3D keypoints for final 6-DoF pose tracking.
Different from the prior work 6-PACK~\cite{wang20206}, where unsupervised keypoint generation may result in a local optimum, we dynamically adapt keypoint generation based on the structural similarity between categorical prior ${P_r}$ and observable point patch ${P^{(t)}}$ in the current frame. In the end, we introduce an auxiliary module to convert ${P^{(t)}}$ into $n$ object key-points $[{k_1}, \ldots ,{k_n}]$, As shown in the right of Fig.~\ref{FIG_overview}, we apply a low-rank Transformer network with $\{ \tilde F_{obj}^{(t)},\tilde F_{temp}^{(t)},\tilde F_{aug}^{(t)}\} $ as query, key and value to estimate a structure regularized projection matrix $M \in {\mathbb{R}^{n \times N}}$ for each category. To encourage the key-point transformation $M$ to adapt the intra-class structural variation among different instances, we utilize the shape prior ${P_r}$ to optimize this auxiliary module by minimizing the loss ${L_{aux}}$ during training step:
\begin{align}
\resizebox{.89\hsize}{!}{$
{L_{aux}} = \sum\limits_{{p_i} \in {P_r}} {\mathop {\min }\limits_{{k_j} \in P_r^K} } ||{p_i} - {k_j}||_2^2 + \sum\limits_{{k_j} \in P_r^K} {\mathop {\min }\limits_{{p_i} \in {P_r}} } ||{p_i} - {k_j}||_2^2,$}
\end{align}
where $P_r^K = M \times {P_r}$ is the prior-base n object keypoints.
This formulation effectively ensures that the 3D space of the key-points is structurally consistent with the shape prior, regardless of the pose change over time.

We then apply a 3D tri-plane as a compact feature representation of the projected keypoints, based on the 2D-base formulation in~\cite{chan2022efficient}. We align these embedding group $\{ \tilde F_{obj}^{(t)},\tilde F_{temp}^{(t)},\tilde F_{aug}^{(t)}\} $ along three axis-aligned orthogonal feature planes by projecting them onto the triplane $\{ {T_{XY}},{T_{YZ}},{T_{XZ}}\}$ using the known camera intrinsics. In our implementation, each plane has dimensions $N \times {d_T}$. For any object key-point, we project it onto each planes, query the corresponding point feature $\{ {T_{xy}},{T_{yx}},{T_{xz}}\} $ via nearest-neighbor point interpolation, which is then conrelated into final keypoint feature $\tilde F_{kp}^{(t)}$.

Due to the identity of the projection matrix $M$, a rough match has been revealed between the keypoint pairs between consecutive frames. Futhermore, we then perfrom the finer keypoint matching to filter possible outlier coarse matches. Following~\cite{sun2021loftr}, a score matrix $S_{kp}$ is calculated based on the simularity between two sets of keypoint features $\tilde F_{kp}^{(t-1)}$ and $\tilde F_{kp}^{(t)}$ in previous and current frames:
\begin{align}
S_{kp}(i,j) = \frac{1}{\tau } \cdot  < \tilde F_{kp}^{(t - 1)}[{k_i}],\tilde F_{kp}^{(t)}[{k_j}] > ,~i,j = 1, \ldots n,
\end{align}
where $\tau $ is a scale factor. We also apply a dual-softmax operator~\cite{tyszkiewicz2020disk} on both dimensions of $S_{kp}$ to obtain the keypoint pairs matching probability:
\begin{align}
{\mathcal{P}_c}(i,j) = softmax{(S_{kp}(i, \cdot ))_j} \cdot softmax{(S_{kp}( \cdot ,j))_i}.
\end{align}
With the confident matrix ${\mathcal{P}_c}$, we select finer key-points with confidence higher than a threshold of ${\theta _c}$, and further enforce the mutual nearest neighbor (MNN) criteria:
\begin{align}
{\mathcal{M}_c} = \{ (i,j)|\forall (i,j) \in MNN({\mathcal{P}_c}),{\mathcal{P}_c}(i,j) \ge {\theta _c}\} .
\end{align}

\subsubsection{Training Supervision}
To improve the performance of our keypoint generation module, we use the following multi-view consistency loss to render the generated keypoints in each of two consecutive frames a better match, placing the keypoint in current view at the transformed corresponding keypoint using ground-truth pose change in previous frame:
\begin{align}
{L_{{\rm{mvc}}}} = \frac{1}{n}\sum\limits_i {||k_i^{(t)} - [\Delta R_{gt}^{(t)}|\Delta T_{gt}^{(t)}] \cdot k_i^{(t - 1)}||},
\end{align}
Meanwhile, to supervise the matching probability matrix ${\mathcal{P}_c}$, we follow LoFTR~\cite{sun2021loftr} to use the negative log-likelihood loss over the grids in $\mathcal{M}_c^{gt}$. We likewise use camera poses and depth maps to compute the ground-truth for ${\mathcal{P}_c}^{gt}$ and $\mathcal{M}_c^{gt}$:  
\begin{align}
{L_{\rm{c}}} =  - \frac{1}{{|\mathcal{M}_c^{gt}|}}\sum\limits_{(i,j) \in \mathcal{M}_c^{gt}} {\log } {\mathcal{P}_c}(i,j).
\end{align}
The above loss functions only guarantees that the generated keypoint pairs are robust to the change in pose. However, it does not ensure these keypoints are optimal for estimating the final pose. In this regard, we use a differentiable pose tracking loss function, which includes a translation loss and a rotation loss:
\begin{align}
{L_{tra}} = ||\frac{1}{n}\sum\limits_i {(k_i^{(t)} - k_i^{(t - 1)}) - \Delta T_{gt}^{(t)}||} ,
\end{align}
\begin{align}
{L_{rot}} = 2\arcsin (\frac{1}{{2\sqrt 2 }}||\Delta {R^{(t)}} - \Delta R_{gt}^{(t)}||).
\end{align}
Therefore, the overall loss function can be determined as the weighted sum of all losses.

\begin{algorithm}[t]
\label{alg}
\SetAlgoLined
\KwIn{Object 6-DoF pose ${{\cal P}^{(t)}} = [{R^{(t)}}|{T^{(t)}}]$ at $t$ time; \\~~~~~~~~~~~The desired object 6-DoF pose ${P^ * } = [{R^ * }|{T^ * }]$.}
\KwOut{Current servo action ${\alpha^{(t)}} \in \{ {\dot \eta ^{(t)}},{v^{(t)}}\}$.}

$//$\emph{Output the low-level action to drive the aerial manipulator}\\
$\Delta R \leftarrow {({R^{(t)}})^T} \cdot {R^ * }$;

$\Delta T \leftarrow {\rm{abs}}({T^ * } - {T^{(t)}})$;

\While{$\Delta R \ge {\delta _R}$ and $\Delta T \ge {\delta _T}$}{
    $//$\emph{Obatin the rotational actions of the manipulator}
    
    $u\theta  \leftarrow \Delta R$;
    
    ${\varepsilon _r} \leftarrow \theta {u^T} - 0$;
     
    Compute Jacobian matrix $L(u,\theta )$ with Eq.~(\ref{jacmatrix});
    
    \eIf{$\theta  \mapsto 0$}{${{\dot \eta }^{(t)}} \leftarrow  - {\lambda _r}J_{mr}^ + {[\begin{array}{*{20}{c}}
{{0_{3 \times 3}}}&{{I_{3 \times 3}}}
\end{array}]^ + }{\varepsilon _r}$\;}{${{\dot \eta }^{(t)}} \leftarrow  - {\lambda _r}J_{mr}^ + {[\begin{array}{*{20}{c}}
{{0_{3 \times 3}}}&{L(u,\theta )}
\end{array}]^ + }{\varepsilon _r}$\;}
    
    $//$\emph{Obatin the translational actions of aerial vehicle} 
    
    $\Delta {m_e} \leftarrow \Delta T$;
    
    ${\varepsilon _p} \leftarrow \Delta {m_e}$;
    
    Compute Jacobian matrix ${J_s}$ with Eq.~(\ref{remove});
    
    ${\upsilon ^{(t)}} \leftarrow  - J_s^ + ({\lambda _p}{L^ + }{\varepsilon _p} + {{\bar J}_s}\nabla )$;
    
   }
   \KwRet{${\alpha^{(t)}}$.}
 \caption{Pose-Aware Discrete Servo Policy (PAD-Servo) for Aerial Manipulator}
\end{algorithm}
\subsection{Pose-Aware Discrete Servo Policy: PAD-Servo}\label{sec:method_b}
\begin{figure}[t]\centering
    \includegraphics[width=\columnwidth]{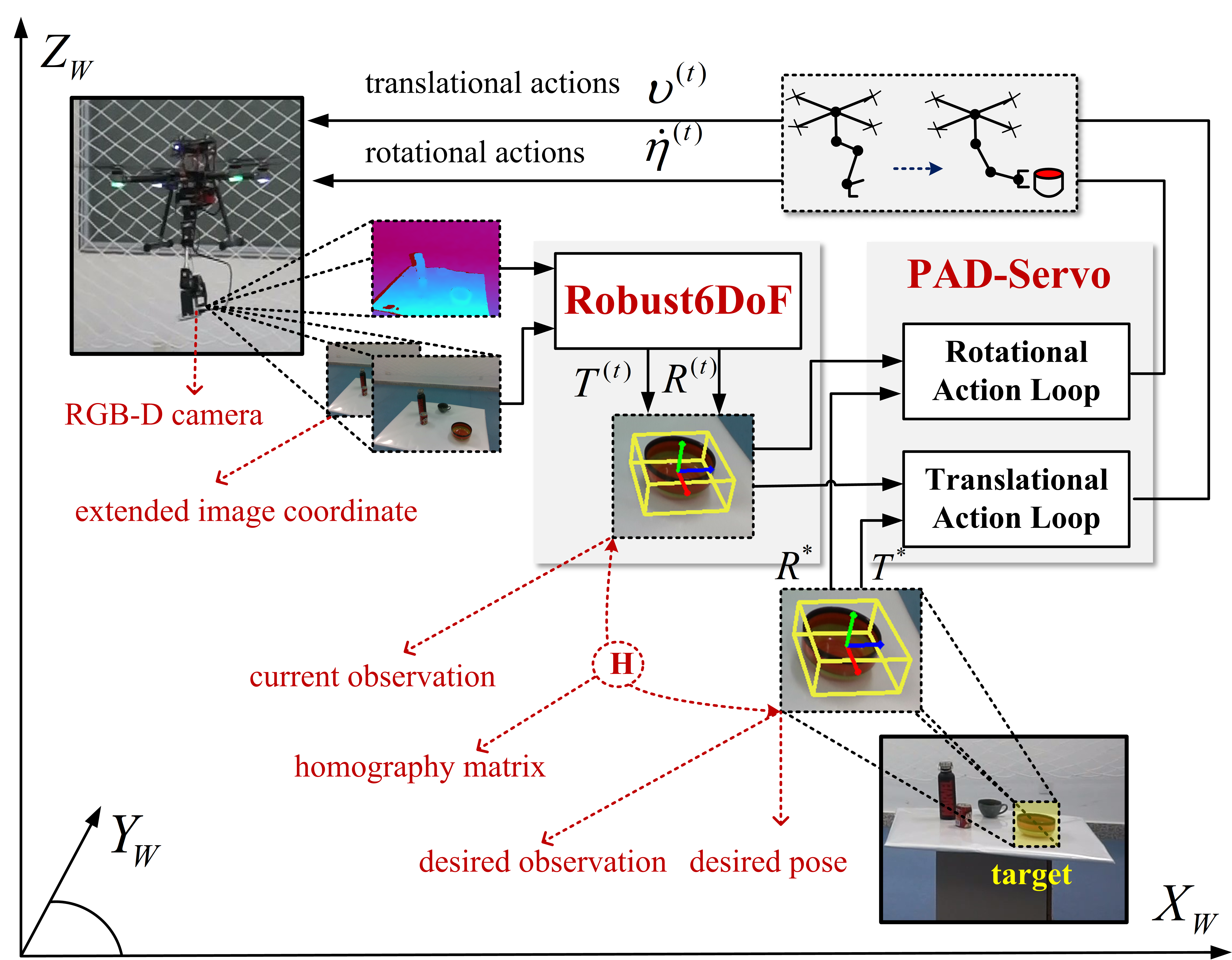}
    \caption{\textbf{Complete flowchart of our proposed PAD-Servo.} According to the object's 6DoF pose ${{\cal P}^{(t)}}$ estimated from our \emph{Robust6DoF} at the current $t$-th timestep, we introduce a decomposed policy to achieve comparable and robust aerial guidance for aerial manipulator.}\label{FIG_policy}\vspace{-0.4cm}
    
\end{figure}

This module is designed to generate the action signals ${\alpha^{(t)}} \in \{ {\dot \eta ^{(t)}},{v^{(t)}}\}$ to accomplish the vision guidance task for aerial manipulator based on the targeted object's 6-DoF pose ${{\cal P}^{(t)}} = [{R^{(t)}}|{T^{(t)}}]$ in the current observation. As depicted in Fig.~\ref{FIG_policy}, we utilize the homography matrix decomposition between the current observation and the desired observation to split the servo process into two parts: the rotational action loop for onboard manipulator and the translational action loop for the aerial vehicle, respectively. Specifically, the rotational action signal is generated from the 3D rotation matrix (${R^{(t)}}$) in 3D Cartesian space, while the translational action signal is derived from the estimated 3D location (${T^{(t)}}$) in 2D image space. For a detailed algorithmic flow, please refer to Alg.~\ref{alg}. The desired observation refers to the image plane where the onboard camera is directly positioned over the targeted object.
It is crucial to emphasize that due to the absence of payload and the minimal sway experienced by the manipulator during the overall guidance process, we primarily focus on the robot's kinematic model, with less emphasis on its dynamic model.
\subsubsection{Rotational Action Loop for Onboard Manipulator}
Given the current estimated 3D rotation ${R^{(t)}}$ and the corresponding desired value ${R^ * }$, we can obtain the change of rotation, \emph{i.e.,}~$\Delta R = {({R^{(t)}})^T} \cdot {R^*}$. Let's use the vector $u\theta $ to express $\Delta R$, where the $u$ represents the rotation axis, and $\theta$ is the rotation angle obtained from identity matrix $\Delta R$.
So the objective function for rotational action can be defined as the error of the $\theta {u^T}$ toward zero, \emph{i.e.,} ${\varepsilon _r} = \theta {u^T}-0$, and its time derivative can be related to the camera velocity component generated from onboard manipulator ${V_C}^{(m)}$:
\begin{align}\label{error}
{\dot \varepsilon _r} = [\begin{array}{*{20}{c}}
0&{{L }(u,\theta )}
\end{array}]{V_C}^{(m)},
\end{align}
where the Jacobian matrix ${L }(u,\theta )$ is
\begin{align}\label{jacmatrix}
\resizebox{.89\hsize}{!}{$
{L }(u,\theta ) = {I_3} - \frac{\theta }{2}{L(u) } + \left( {1 - {{\sin c(\theta )} \mathord{\left/
 {\vphantom {{\sin c(\theta )} {\sin {c^2}(\frac{\theta }{2})}}} \right.
 \kern-\nulldelimiterspace} {\sin {c^2}(\frac{\theta }{2})}}} \right)L(u) ^{\rm{2}},$}
\end{align}
$\sin c(\theta ) = {{\sin (\theta )} \mathord{\left/
 {\vphantom {{\sin (\theta )} \theta }} \right.
 \kern-\nulldelimiterspace} \theta }$ and ${L(u)}$ is antisymmetric matrix associated with $u$. The error ${\varepsilon _r}$ can be converged exponentially by imposing ${\dot \varepsilon _r} = {-\lambda _r}{\varepsilon _r}$ and $\lambda _r$ tunes the convergence rate.
Meanwhile, based on the special form of ${L }(u,\theta )$, we can set ${L }(u,\theta ) = {L }(u,\theta ) ^{ - 1} = I_{3 \times 3}$ for the small value of $\theta$. 
Then we compute the relationship between the vector ${V_C}^{(m)}$ and the angular velocity vector of each joint of the manipulator $\dot \eta$, which can be expressed as:
\begin{align}\label{relation}
{V_C}^{(m)} &= \left[ {\begin{array}{*{20}{c}}
{R_B^C}&0\\
0&{R_B^C}
\end{array}} \right]{J_m}{\left[ {\begin{array}{*{20}{c}}
{{{\dot \eta }_1}}&{{{\dot \eta }_2}}&{{{\dot \eta }_3}}&{{{\dot \eta }_4}}
\end{array}} \right]^T} \notag\\&= \bar R_B^C{J_m}{\dot \eta ^T},
\end{align}
where $J_m$ is the arm Jacobian matrix, and ${R_B^C}$ is the rotation matrix of the base frame with respect to the camera frame. Finally, according to the Eq.~(\ref{error}) and (\ref{relation}), the rotational servo action law of the manipulator can be described as:
\begin{align}\label{joint-signs}
\resizebox{.89\hsize}{!}{$
{\dot \eta ^{(t)}} = \left\{ {\begin{array}{*{20}{c}}
{ - {\lambda _r}J_{mr}^ + {{[\begin{array}{*{20}{c}}
{{0_{3 \times 3}}}&{{I_{3 \times 3}}}
\end{array}]}^ + }{\varepsilon _r}}\\
{ - {\lambda _r}J_{mr}^ + {{[\begin{array}{*{20}{c}}
{{0_{3 \times 3}}}&{L(u,\theta )}
\end{array}]}^ + }{\varepsilon _r}}
\end{array}\begin{array}{*{20}{c}}
{}\\
{}
\end{array}\begin{array}{*{20}{c}}
{if\theta  \to 0}\\
{otherwise}
\end{array}} \right.,$}
\end{align}
where ${J_{mr}} = \bar R_B^C{J_m} \in {\mathbb{R}^{6 \times 4}}$.
\subsubsection{Translational Action Loop for Aerial Vehicle}
For the translational action for aerial vehicle, we can define the corresponding objective function as the ${m_e}$ toward the desired value $m_e^*$, \emph{i.e.,} ${\varepsilon _p} = {({m_e} - m_e^*)^T}$, where ${m_e}$ is the extended image coordinate:
\begin{align}
{m_e} = {[\begin{array}{*{20}{c}}
x&y&z
\end{array}]^T} = {[\begin{array}{*{20}{c}}
{{X \mathord{\left/
 {\vphantom {X Z}} \right.
 \kern-\nulldelimiterspace} Z}}&{{Y \mathord{\left/
 {\vphantom {Y Z}} \right.
 \kern-\nulldelimiterspace} Z}}&{\log Z}
\end{array}]^T},
\end{align}
where ${T^{(t)}} = {[\begin{array}{*{20}{c}}
X&Y&Z
\end{array}]^T}$ is 3D location of targeted object in the current observation. Similarly, the time derivate of this error function can be related to the camera velocity componemt from aerial vehicle ${V_C}^{(a)}$:
\begin{align}
{{\dot \varepsilon }_p} = [\begin{array}{*{20}{c}}
{{L_Z}}&{{L(x,y)}}
\end{array}]{V_C}^{(a)} = L{V_C}^{(a)},
\end{align}
where ${L(x,y)}$ is:
\begin{align}
{L(x,y)} = \left[ {\begin{array}{*{20}{c}}
{xy}&{ - (1 + {x^2})}&y\\
{\left( {1 + {y^2}} \right)}&{ - xy}&{ - x}\\
{ - y}&x&0
\end{array}} \right],
\end{align}
and the upper triangular matrix ${L_Z }$ is given by:
\begin{align}
{L_Z } = \frac{1}{{Z}}\left[ {\begin{array}{*{20}{c}}
{{\rm{ - }}1}&0&x\\
0&{{\rm{ - }}1}&y\\
0&0&{{\rm{ - }}1}
\end{array}} \right],
\end{align}
where the matrix ${L_Z }$ is obtained from decomposing the homograph matrix between current and desired observations. For additional detailed content about the homography decomposition, we refer the reader to \cite{malis19992}.

Similar to the ${\varepsilon _r}$, we also impose ${\varepsilon _p}$ to the exponential convergence, \emph{i.e.,}
${{\dot \varepsilon }_p} =  - {\lambda _p}{\varepsilon _p}$ by setting the convergence rate ${\lambda _p}$. In this way, such camera velocity component ${V_C}^{(a)}$ can be expressed as:
\begin{align}\label{velocity com}
{V_C}^{(a)} =  - {\lambda _p}{L^ + }{\varepsilon _p},
\end{align}
where ${L^ + }$ is Moore-Penrose matrix of ${L}$. Meanwhile, the relationship between the velocities of aerial vehicle and the camera velocity component from aerial vehicle ${V_C}^{(a)}$ can be expressed as:
\begin{align}
\resizebox{.89\hsize}{!}{$
\begin{array}{l}
{V_C}^{(a)}{\rm{ = }}\left[ {\begin{array}{*{20}{c}}
{R_B^C}&{ - R_B^Cr_C^B}\\
0&{R_B^C}
\end{array}} \right]{[\begin{array}{*{20}{l}}
{{\dot p^T},{\omega ^T}}
\end{array}]^T}{\rm{ }} {\rm{ = }}\bar r_C^B{ \left[ {{\dot p^T},{\omega ^T}} \right] }\mathop{{}}\nolimits^{{T}},
\end{array}$}
\end{align}
where ${r_C^B}$ is the distance vector between the base frame and the camera frame. According to the underactuation of aerial vehicle, we remove the uncontrollable variables $\nabla  = {({\omega _x},{\omega _y})^T}$ from the translational and angular velocity vector of the aerial vehicle:
\begin{align}\label{remove}
{V_C}^{(a)} = {J_s}\upsilon  + {{\bar J}_s}\nabla ,
\end{align}
where ${{\bar J}_s}$ is the Jacobian formed by the columns of $\bar r_C^B$ corresponding to ${{\omega _x}}$ and ${{\omega _y}}$, and ${J_s}$ is the Jacobian formed by all other columns of $\bar r_C^B$ corresponding to $\upsilon  = ({v_x},{v_y},{v_z},{\omega _z})^T$.
According to Eq.~(\ref{velocity com}) and~(\ref{remove}), the translational servo action law of aerial vehicle can be formulated as:
\begin{align}\label{vehicle sign}
{\upsilon ^{(t)}}  =  - J_s^ + ({\lambda _p}{L^ + }{\varepsilon _p} + {{\bar J}_s}\nabla ).
\end{align}

\section{Experiments}\label{sec:experiment}

In this section, we first present extensive quantitative comparative experiments on the four widely-used public datasets to evaluate the performance of the presented category-level 6-DoF pose tracker \emph{Robust6DoF} and compare it with currently available state-of-the-art baselines. We also perform numerous ablation studies and additional analyses to verify the advantages of each component in our method. In addition, to further test the effectiveness of the proposed completed pipline, 
we implement a visual guidance experiment directly using our model trained on the public dataset, along with the proposed \emph{PAD-Servo}, to control a real-world aerial robot platform, namely, an aerial manipulator in our Robotic Laboratory.

\subsection{Experimental Setup}

\subsubsection{Datasets}
We evaluate \emph{Robust6DoF} using four public datasets, \emph{i.e.,} NOCS-REAL275~\cite{wang2019normalized}, YCB-Video~\cite{xiang2017posecnn}, YCBInEOAT~\cite{wen2020se} and Wild6D~\cite{ze2022category} dataset. The NOCS-REAL275 dataset was proposed by Wang~\cite{wang2019normalized}and  contains six categories: \emph{bottle, bowl, camera, can, laptop and mug}. It includes 13 real-world scenes, with 7 scenes (4.3K RGB-D images) for training and 6 scenes (2.7K RGB-D images) for testing. The training and testing sets include 18 real object instances across these six categories. The YCB-Video dataset was introduced in~\cite{xiang2017posecnn} and consists of both real-world and synthetic images. We use only its real-world data for training, which includes 92 videos captured in various settings using an RGB-D camera. During training, we utilize $80$ of these videos, reserving the remaining $12$ for testing. 
The YCBInEOAT dataset~\cite{wen2020se} considers five YCB-Video objects, including \emph{mustard bottle, tomato soup can, sugar box, bleach cleanser and cracker box}. It contains 9 video sequences captured by a static RGB-D camera.
The Wild6D~\cite{ze2022category} is a large-scale RGB-D dataset that consists of $5,166$ videos (over $1.1$ million images) featuring $1722$ different object instances across five categories: \emph{bottle, bowl, camera, laptop, and mug.} Following the creator's instuctions, we treat 486 videos of 162 instances as the test set.
\subsubsection{Evaluation Metrics}
We use the following four types of evaluation metrics: 
\begin{itemize}
    \item~${IoUx}$. It measures the average percision for various IoU-overlap thresholds, which calculates the overlap between two 3D bounding boxes based on the predicted pose and the ground-truth pose. 
    \item ~\textbf{${a^ \circ }b\,cm$}. It quantifies the pose estimation error for rotation and translation, and the error is less than $a^ \circ$ for rotation and $b~cm$ for translation. We adopt the ${5^ \circ }2~cm$, ${5^ \circ }5~cm$, ${10^ \circ }2~cm$ and ${10^ \circ }5~cm$ for evaluation. 
    \item~\textbf{${\rm{ADD (S)}}$}. Evaluating for instance-level 6-DoF pose tracking. ADD measures the distance between the ground truth 3D model and the posed model using predictions. ADD-S is for the symmetrical object. 
    \item~\textbf{${R_\mathit{err}}$}~(\textbf{${T_\mathit{err}}$}). These terms measure the average error of rotation (degrees) and translation (centimeters), that are used for category-level pose tracking. 
\end{itemize}

\subsubsection{Implementation Details}
All the building blocks in the \emph{Robust6DoF}'s network are trained using an ADAM optimizer with an initial learning rate of ${10^{ - 3}}$ and a batch size of $32$. The training epoch number is set as $50$. 
The experiments on the public datasets were conducted on a desktop computer with an Intel Xeon Gold 6226R@2.90GHz processor and a single NVIDIA RTX A6000 GPU. We trained our model using the NOCS-REAL275 dataset and fine-tuned it on the YCB-Video dataset.
The number of the partially visible point patch, ${P^{(t)}}$ and the priori shape-point ${P_r}$ are both set to $N = {N_r} = 2048$.
The number of generated key-points is $n = 512$.
The confidence threshold is set to ${\theta _c} = 0.45$.
In real-world experiment, we implement Mask-RCNN for segmentation, as in~\cite{wang2019normalized}.
The gains of our \emph{PAD-Servo} in Eq.~(\ref{joint-signs}) and Eq.~(\ref{vehicle sign}) are empirically set as follows: ${\lambda _r} = 0.25$, ${\lambda _p} = 0.27$ and the guidance end thresholds are set to ${\delta _R} = 0.075$, ${\delta _T} = 0.040$.
\newpage
\begin{table*}[ht]
\scriptsize
\footnotesize
\setlength{\tabcolsep}{3.0pt}
\centering
\renewcommand\arraystretch{1.0}
\caption{\textbf{Quantitative comparison of category-level 6-DoF pose estimation on the pubilc NOCS-REAL275 dataset.} Note that the best and the second best results are highlighted in \textbf{bold} and \underline{underlined}. The results are averaged over all six categories. The comparison results of current state-of-the-art baselines are all summarized from their original papers and empty denotes no results are reported under their original paper.}

\label{table_nocs-estimation}
\vspace{-0.2cm}
\begin{tabular}{l|cc|ccccccc|c}
\toprule [0.8pt]
\multirow{2}{*}{Method} & \multirow{2}{*}{\begin{tabular}[c]{@{}c@{}}Training Data\end{tabular}} & \multirow{2}{*}{\begin{tabular}[c]{@{}c@{}}Shape Prior\end{tabular}} & \multicolumn{7}{c|}{Evaluation Metrics}                                                 & \multirow{2}{*}{\begin{tabular}[c]{@{}c@{}}\#Params\\ (M)($\downarrow$)\end{tabular}} \\ 
                        &                                                                          &                                                                        & $IoU25\uparrow$ & $IoU50\uparrow$ & \multicolumn{1}{c|}{$IoU75\uparrow$} & ${5^ \circ }2cm \uparrow $ & ${5^ \circ }5cm \uparrow $ & ${10^ \circ }2cm \uparrow $ & ${10^ \circ }5cm \uparrow $ &                                                                            \\ \midrule
NOCS~\cite{wang2019normalized}~{\color{gray}[CVPR2019]}                    & RGB                                                                      & \usym{2717}                                                                      & \underline{84.9}  & 80.5  & \multicolumn{1}{c|}{30.1}  & 7.2   & 10    & 13.8   & 25.2   & -                                                                          \\
SPD~\cite{tian2020shape}~{\color{gray}[ECCV2020]}                     & RGB-D                                                                    & \usym{2713}                                                                      & 83.4  & 77.3  & \multicolumn{1}{c|}{53.2}  & 19.3  & 21.4  & 43.2   & 54.1   & 18.3                                                                       \\
SGPA~\cite{chen2021sgpa}~{\color{gray}[ICCV2021]}                    & RGB-D                                                                    & \usym{2713}                                                                      & -     & 80.1  & \multicolumn{1}{c|}{61.9}  & 35.9  & 39.6  & 61.3   & 70.7   & 23.3                                                                       \\
CR-Net~\cite{wang2021category}~{\color{gray}[IROS2021]}                  & RGB-D                                                                    & \usym{2713}                                                                      & -     & 79.3  & \multicolumn{1}{c|}{55.9}  & 27.8  & 34.3  & 47.2   & 60.8   & 21.4                                                                       \\
CenterSnap~\cite{irshad2022centersnap}~{\color{gray}[ICRA2022]}              & RGB-D                                                                    & \usym{2717}                                                                      & -     & 80.2  & \multicolumn{1}{c|}{-}     & -     & 27.2  & -      & 58.8   & -                                                                          \\
ShAPO~\cite{irshad2022shapo}~{\color{gray}[ECCV2022]}                   & RGB-D                                                                    & \usym{2717}                                                                      & -     & 79.0  & \multicolumn{1}{c|}{-}     & -     & 48.8  & -      & 66.8   & -                                                                          \\
TTA-COPE~\cite{lee2023tta}~{\color{gray}[CVPR2023]}                & RGB-D                                                                    & \usym{2717}                                                                      & -     & 69.1  & \multicolumn{1}{c|}{39.7}  & 30.2  & 35.9  & 61.7   & 73.2   & -                                                                          \\
IST-Net~\cite{liu2023prior}~{\color{gray}[ICCV2023]}                 & RGB-D                                                                    & \usym{2717}                                                                      & 84.3  & 82.5  & \multicolumn{1}{c|}{\underline{76.6}}  & 47.5  & 53.4  & 72.1   & 80.5   & -                                                                          \\ \midrule
FS-Net~\cite{chen2021fs}~{\color{gray}[CVPR2021]}                  & D                                                                        & \usym{2717}                                                                      & 84.0  & 81.1  & \multicolumn{1}{c|}{63.5}  & 19.9  & 33.9  & -      & 69.1   & 41.2                                                                       \\
UDA-COPE~\cite{lee2022uda}~{\color{gray}[CVPR2022]}                & D                                                                        & \usym{2717}                                                                      & -     & 79.6  & \multicolumn{1}{c|}{57.8}  & 21.2  & 29.1  & 48.7   & 65.9   & -                                                                          \\
SAR-Net~\cite{lin2022sar}~{\color{gray}[CVPR2022]}                 & D                                                                        & \usym{2717}                                                                      & -     & 79.3  & \multicolumn{1}{c|}{62.4}  & 31.6  & 42.3  & 50.3   & 68.3   & \underline{6.3}                                                                        \\
GPV-Pose~\cite{di2022gpv}~{\color{gray}[CVPR2022]}                 & D                                                                        & \usym{2717}                                                                      & 84.1  & \underline{83.0}  & \multicolumn{1}{c|}{64.4}  & 32.0  & 42.9  & 55.0   & 73.3   & 8.6                                                                        \\
HS-Pose~\cite{zheng2023hs}~{\color{gray}[CVPR2023]}                 & D                                                                        & \usym{2717}                                                                      & 84.2  & 82.1  & \multicolumn{1}{c|}{74.7}  & 46.5  & 55.2  & 68.6   & 82.7   & -                                                                          \\
Query6DoF~\cite{wang2023query6dof}~{\color{gray}[ICCV2023]}               & D                                                                        & \usym{2713}                                                                      & -     & 82.5  & \multicolumn{1}{c|}{76.1}  & \underline{49.0}  & \underline{58.9}  & \underline{68.7}   & \underline{83.0}   & -                                                                          \\
GPT-COPE~\cite{zou2023gpt}~{\color{gray}[TCSVT2023]}                & D                                                                        & \usym{2713}                                                                      & -     & 82.0  & \multicolumn{1}{c|}{70.4}  & 45.9  & 53.8  & 63.1   & 77.7   & 7.1                                                                        \\ \midrule 
Ours                    & RGB-D                                                                        & \usym{2713}                                                                      & \textbf{89.8}  & \textbf{87.0}  & \multicolumn{1}{c|}{\textbf{82.5}}  & \textbf{57.1}  & \textbf{70.6}  & \textbf{75.2}   & \textbf{84.5}   & \textbf{6.0}                                                                        \\ \bottomrule [0.8pt]
\end{tabular}
\vspace{-0.2cm}
\end{table*}

\begin{table*}[ht]
\scriptsize
\footnotesize
\setlength{\tabcolsep}{4.5pt}
\centering
\renewcommand\arraystretch{1.0}
\caption{\textbf{Quantitative comparison of category-level 6-DoF pose tracking on the pubilc NOCS-REAL275 dataset.} Note that the best and the second best results are highlighted in \textbf{bold} and \underline{underlined}. The results of available baselines are all summarized from their original papers.}
\label{table_nocs-tracking}
\vspace{-0.2cm}
\begin{tabular}{c|c|c|c|c|c|c|c|c}
\toprule [0.8pt]
Method         & ICP~\cite{wang20206}   & Oracle ICP~\cite{weng2021captra} & 6-PACK~\cite{wang20206} & \begin{tabular}[c]{@{}c@{}}6-PACK \\ w/o temporal~\cite{wang20206}\end{tabular} & CAPTRA~\cite{weng2021captra} & \begin{tabular}[c]{@{}c@{}}CAPTRA\\ +RGB seg.~\cite{weng2021captra}\end{tabular} & MaskFusion~\cite{runz2018maskfusion}  & Ours     \\ \midrule
Input                               & Depth & Depth      & RGB-D  & RGB-D                         & Depth  & RGB-D                     & RGB-D        & RGB-D \\ \midrule
Initialization                      & GT.   & GT.        & GT.    & Pert.                         & Pert.  & Pert.                     & GT.        & GT.       \\ \midrule
${5^ \circ }5cm \uparrow $          & 16.9  & 0.65       & 28.9   & 22.1                          & 62.2   & \underline{63.6}             & 26.5       & \textbf{70.6}      \\
$IoU25\uparrow$                     & 47.0  & 14.7       & 55.4   & 53.6                          & 64.1   & \underline{69.2}          & 64.9       & \textbf{89.8}      \\
${R_{err}}~\downarrow$              & 48.1  & 40.3       & 19.3   & 19.7                          & \underline{5.9}    & 6.4           & 28.5       & \textbf{5.2}      \\
${T_{err}}~\downarrow$              & 10.5  & 7.7        & \textbf{3.3}    & \underline{3.6}      & 7.9    & 4.2                       & 8.3        & \textbf{3.0}      \\ \bottomrule [0.8pt]
\end{tabular}
\vspace{-0.2cm}
\end{table*}

\subsection{Quantitative Comparisons on the Public Datasets}
\subsubsection{Results on the NOCS-REAL275 Dataset}
\begin{figure}[t]
    \centering
    \includegraphics[width=0.49\textwidth]{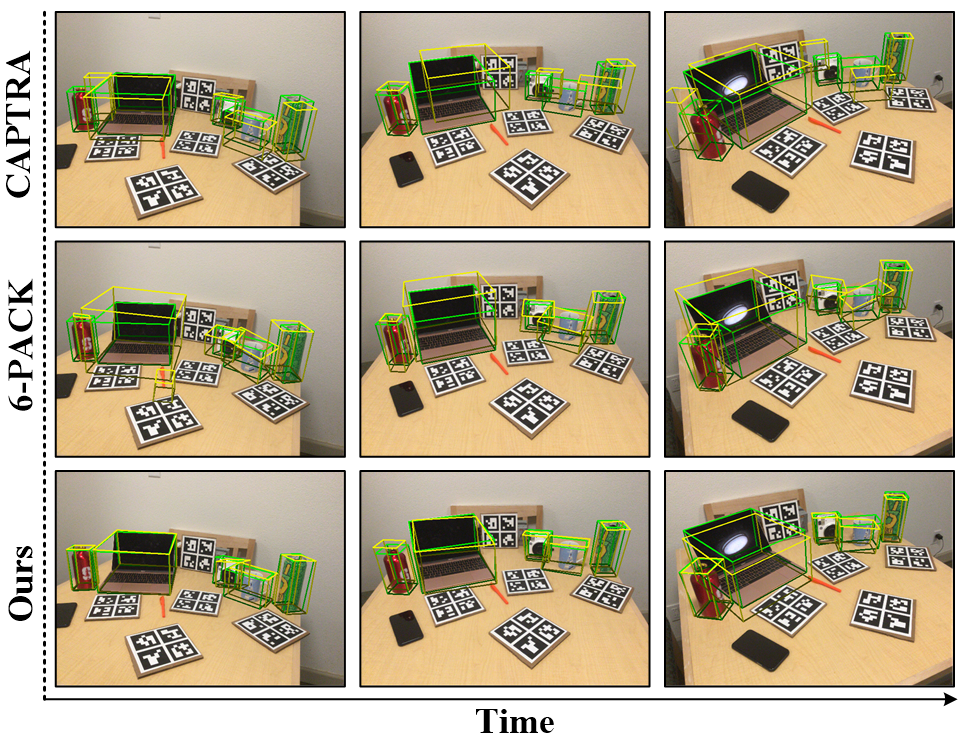} 
    \vspace{-0.4cm}
    \caption{\textbf{Visualization comparison on NOCS-REAL275 dataset.} We compare \emph{Robust6DoF} with representative category-level 6-DoF pose tracking methods (6-PACK~\cite{wang20206} and CAPTRA~\cite{weng2021captra}) . Yellow and green represent the results from prediction and ground-truth label.}\label{FIG_show_nocs}
    \vspace{-0.2cm}
\end{figure}

We first conduct both category-level 6-DoF pose tracking and estimation on the testing set of the NOCS-REAL275 dataset. Some quantitative results are presented in TABLE~\ref{table_nocs-estimation}, TABLE~\ref{table_nocs-tracking} and Fig.~\ref{FIG_show_nocs}. As shown in TABLE~\ref{table_nocs-estimation}, we compare our approach with $15$ state-of-the-art single estimation-based methods. These baselines either take RGB~(-D) as inputs or use only point cloud features (D), and they can be divided into two groups: shape prior-based and prior-free methods. In detail, we outperform the pioneer work NOCS~\cite{wang2019normalized} by 52.4 in $IoU75$, 49.9 in ${5^ \circ }2cm$ and 60.6 in ${5^ \circ }5cm$. For comparison with prior-free methods, we also achieve better results than existing approaches. In particular, we outperform Query6DoF~\cite{wang2023query6dof}, the current most powerful method, by 57.1 vs 49.0 on ${5^ \circ }2cm$, 70.6 vs 58.9 on ${5^ \circ }5cm$ and 75.2 vs 68.7 on ${10^ \circ }2cm$. As for prior-based methods, we also show significant improvements in nearly all the evaluation metrics with large margins. For example, we reach 87.0, 82.5, and 84.5 in terms of $IoU50$, $IoU75$ and ${10^ \circ }5cm$, which outperform the most competitive representative work~SGPA~\cite{chen2021sgpa} by 6.9\%, 20.6\% and 13.8\%. Notably, our model has minimal parameters among all baselines, proving its low computing cost.

In addition, we summarize the quantitative results for category-level object 6-DoF pose tracking, as depicted in TABLE~\ref{table_nocs-tracking}. We compare our method with the currently available state-of-the-art tracking methods: classic ICP~\cite{weng2021captra} approach and its improved version OracleICP~\cite{weng2021captra}, 6-PACK~\cite{wang20206} and CAPTRA~\cite{weng2021captra} along with their variants, and MaskFusion~\cite{runz2018maskfusion}. It is worth noting that our method also achieves the best performance in terms of all track-based evaluation metrics. The corresponding quantitative comparisons are presented in Fig.~\ref{FIG_show_nocs}, which are arranged from left to right in time sequence. It further shows that our tracking results more accurately match the ground truth compared to CAPTRA~\cite{weng2021captra} and 6-PACK~\cite{wang20206}.

\begin{table}[ht]
\scriptsize
\footnotesize
\setlength{\tabcolsep}{2.4pt}
\centering
\renewcommand\arraystretch{1.1}
\caption{\textbf{Quantitative comparison of instance-level 6-DoF pose tracking on the pubilc YCB-Video dataset.} Note that the best and the second best results are highlighted in \textbf{bold} and \underline{underlined}, respectively.}
\label{table_ycb_video_compare}
\vspace{-0.2cm}
\begin{tabular}{l|cc|cc|cc}
\toprule [0.8pt]
\multicolumn{1}{c|}{\multirow{2}{*}{Objects}} & \multicolumn{2}{c|}{PoseCNN~\cite{xiang2017posecnn}} & \multicolumn{2}{c|}{CatTrack~\cite{yu2023cattrack}} & \multicolumn{2}{c}{Ours} \\ \cmidrule{2-7} 
\multicolumn{1}{c|}{}                         & ADD          & ADD-S         & ADD           & ADD-S         & ADD        & ADD-S       \\ \midrule
002 master chef can                           & 50.9         & 84.0          & \underline{82.5}          & \underline{86.3}          & \textbf{90.3}       & \textbf{91.2}        \\
    003 cracker box                               & 51.7         & \underline{76.9}          & \underline{86.2}          & \textbf{91.7}          & \textbf{92.1}       & \textbf{91.7}        \\
004 sugar box                                 & 68.6         & \underline{84.3}          & 83.6          & \textbf{92.0}          & \textbf{88.5}       & \underline{89.2}        \\
005 tomato soup can                           & 66.0         & 80.9          & \underline{84.3}          & \underline{88.6}          & \textbf{86.5}       & \textbf{90.3}        \\
006 mustard bottle                            & 79.9         & \underline{90.2}          & \underline{85.9}          & \underline{90.2}          & \textbf{91.4}       & \textbf{90.3}        \\
007 tuna fish can                             & 70.4         & 87.9          & \underline{84.7}          & \textbf{91.5}          & \textbf{88.7}       & \underline{89.2}        \\
008 pudding box                               & \textbf{92.9}         & 79.0          & 73.4          & \underline{85.8}          & \underline{82.5}       & \textbf{86.4}        \\
009 gelatin box                               & 75.2         & 87.1          & \underline{90.8}          & \textbf{93.9}          & \textbf{90.9}       & \underline{91.4}        \\
010 potted meat can                           & 59.6         & \underline{78.5}          & \underline{66.7}          & 75.9          & \textbf{80.1}       & \textbf{82.5}        \\
011 banana                                    & 72.3         & \textbf{85.9}          & \underline{76.8}          & 82.4          & \textbf{81.2}       & \underline{83.5}        \\
019 pitcher base                              & 52.5         & 76.8          & \underline{84.1}          & \textbf{92.8}          & \textbf{86.4}       & \underline{88.8}        \\
021 bleach cleanser                           & 50.5         & 71.9          & \underline{73.4}          & \underline{80.5}          & \textbf{79.8}       & \textbf{80.9}        \\
024 bowl                                      & 6.5          & 69.7          & \underline{33.6}          & \textbf{89.8}          & \textbf{85.4}       & \underline{86.5}        \\
025 mug                                       & 57.7         & \textbf{78.0}          & 72.1          & \textbf{83.9}          & \underline{77.2}       & \underline{80.4}        \\
035 power drill                               & 55.1         & 72.8          & \underline{71.3}          & \textbf{86.0}          & \textbf{79.5}       & \underline{82.6}        \\
036 wood block                                & \underline{31.8}         & \underline{65.8}          & 28.6          & 62.3          & \textbf{76.8}       & \textbf{81.3}        \\
037 scissors                                  & 35.8         & 56.2          & \underline{64.9}          & \underline{74.3}          & \textbf{74.9}       & \textbf{80.0}        \\
040 large marker                              & 58.0         & 71.4          & \underline{70.8}          & \textbf{83.4}          & \textbf{81.0}       & \underline{82.7}        \\
051 large clamp                               & 25.0         & 49.9          & \underline{66.8}          & \underline{78.1}          & \textbf{76.9}       & \textbf{79.4}        \\
052 extra large clamp                         & 15.8         & 47.0          & \underline{49.8}          & \underline{77.2}          & \textbf{72.1}       & \textbf{77.7}        \\
061 foam brick                                & 40.4         & 87.8          & \underline{86.0}          & \textbf{93.4}          & \textbf{89.9}       & \underline{92.1}        \\ \midrule
\multicolumn{1}{c|}{Average}                  & 53.7         & 75.9          & \underline{72.2}          & \underline{84.8}          & \textbf{83.4}       & \textbf{85.6}        \\ \bottomrule [0.8pt]
\end{tabular}
\vspace{-0.2cm}
\end{table}
\begin{figure}[t]
    \centering
    \includegraphics[width=0.49\textwidth]{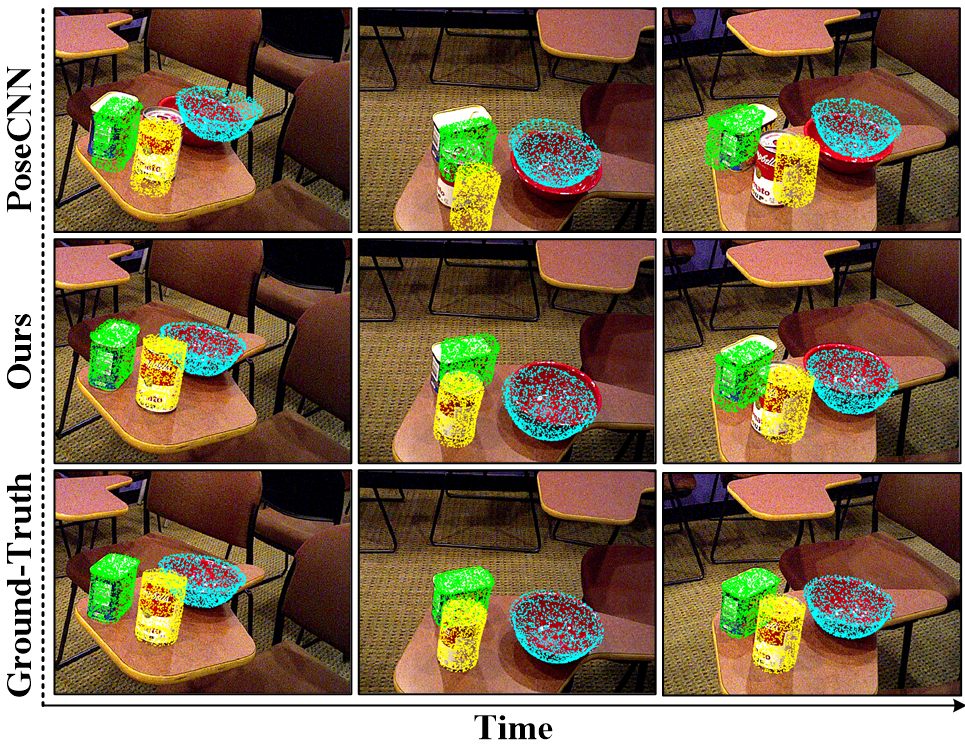} 
    \vspace{-0.4cm}
    \caption{\textbf{Visualization comparison on YCB-Video dataset.} We compare \emph{Robust6DoF} with representative instance-level baseline (PoseCNN~\cite{xiang2017posecnn}). To keep in line with PoseCNN, each object shape model are transformed with the predicted pose and then projected into the 2D images.}\label{FIG_show_ycb-video}
    \vspace{-0.2cm}
\end{figure}

\begin{table}[]
\scriptsize
\footnotesize
\setlength{\tabcolsep}{1.5pt}
\centering
\renewcommand\arraystretch{1.1}
\caption{\textbf{Quantitative comparison on public YCBInEOAT dataset.} We measure using ADD and ADD-S metrics.Note that the best and the second best results are highlighted in \textbf{bold} and \underline{underlined}, respectively. The results of RGF~\cite{issac2016depth}, POT~\cite{wuthrich2013probabilistic}, MaskFusion~\cite{runz2018maskfusion} and TEASER~\cite{yang2020teaser} are all summarized from the literature~\cite{wen2021bundletrack}.}
\label{table_ycbineoat_compare}
\vspace{-0.2cm}
\begin{tabular}{ll|cc|ccc}
\toprule [0.8pt]
\multicolumn{2}{c|}{Method}                                       & \begin{tabular}[c]{@{}c@{}}RGF \\ \cite{issac2016depth}\end{tabular} & \begin{tabular}[c]{@{}c@{}}POT\\ \cite{wuthrich2013probabilistic}\end{tabular} & \begin{tabular}[c]{@{}c@{}}MaskFusion\\ \cite{runz2018maskfusion}\end{tabular} & \multicolumn{1}{c|}{\begin{tabular}[c]{@{}c@{}}TEASER\\ \cite{yang2020teaser}\end{tabular}} & Ours  \\ \midrule
\multicolumn{2}{c|}{Setting}                                      & \multicolumn{2}{c|}{3D Model}                                                                                       & \multicolumn{3}{c}{No Model}                                                                                                                             \\ \midrule
\multicolumn{1}{l|}{\multirow{2}{*}{003 cracker box}}     & ADD   & 34.78                                                   & 79.00                                                     & \underline{79.74}                                                         & \multicolumn{1}{c|}{63.24}                                                       & \textbf{80.51} \\
\multicolumn{1}{l|}{}                                     & ADD-S & 55.44                                                   & 88.13                                                     & \textbf{88.28}                                                         & \multicolumn{1}{c|}{81.35}                                                       & \underline{86.32} \\ \hline
\multicolumn{1}{l|}{\multirow{2}{*}{021 bleach cleanser}} & ADD   & 29.40                                                   & 61.47                                                     & 29.83                                                         & \multicolumn{1}{c|}{\underline{61.83}}                                                       & \textbf{79.25} \\
\multicolumn{1}{l|}{}                                     & ADD-S & 45.03                                                   & 68.96                                                     & 43.31                                                         & \multicolumn{1}{c|}{\underline{82.45}}                                                       & \textbf{83.19} \\ \hline
\multicolumn{1}{l|}{\multirow{2}{*}{004 sugar box}}       & ADD   & 15.82                                                   & \underline{86.78}                                                     & 36.18                                                         & \multicolumn{1}{c|}{51.91}                                                       & \textbf{89.11} \\
\multicolumn{1}{l|}{}                                     & ADD-S & 16.87                                                   & \underline{92.75}                                                     & 45.62                                                         & \multicolumn{1}{c|}{81.42}                                                       & \textbf{94.42} \\ \hline
\multicolumn{1}{l|}{\multirow{2}{*}{005 tomato soup can}} & ADD   & 15.13                                                   & \underline{63.71}                                                     & 5.65                                                          & \multicolumn{1}{c|}{41.36}                                                       & \textbf{85.77} \\
\multicolumn{1}{l|}{}                                     & ADD-S & 26.44                                                   & \textbf{93.17}                                                     & 6.45                                                          & \multicolumn{1}{c|}{71.61}                                                       & \underline{90.79} \\ \hline
\multicolumn{1}{l|}{\multirow{2}{*}{006 mustard bottle}}  & ADD   & 56.49                                                   & \underline{91.31}                                                     & 11.55                                                         & \multicolumn{1}{c|}{71.92}                                                       & \textbf{92.23} \\
\multicolumn{1}{l|}{}                                     & ADD-S & 60.17                                                   & \underline{95.31}                                                     & 13.11                                                         & \multicolumn{1}{c|}{88.53}                                                       & \textbf{96.36} \\ \midrule
\multicolumn{1}{l|}{\multirow{2}{*}{Average}}             & ADD   & 29.98                                                   & \underline{78.28}                                                     & 35.07                                                         & \multicolumn{1}{c|}{57.91}                                                       & \textbf{85.37} \\
\multicolumn{1}{l|}{}                                     & ADD-S & 39.90                                                   & \underline{89.18}                                                     & 41.88                                                         & \multicolumn{1}{c|}{81.17}                                                       & \textbf{90.22} \\ \bottomrule [0.8pt]
\end{tabular}
\vspace{-0.2cm}
\end{table}

\begin{table}[h]

\scriptsize
\footnotesize
\setlength{\tabcolsep}{2.0pt}
\centering
\renewcommand\arraystretch{1.0}
\caption{\textbf{Quantitative comparison on the pubilc Wild6D dataset.} The results of state-of-the-arts are summarized from~\cite{ze2022category}. Note that the best and the second best results are highlighted in \textbf{bold} and \underline{underlined}.}
\label{table_wild6d}
\vspace{-0.2cm}
\begin{tabular}{l|c|cccc}
\toprule [0.8pt]
\multirow{2}{*}{Method} & \multirow{2}{*}{Prior} & \multicolumn{4}{c}{Evaluation Metrics}    \\  
                        &                              & $IoU50$ & ${5^ \circ }2cm$ & ${5^ \circ }5cm$ & ${10^ \circ }5cm$ \\ \midrule
SPD~\cite{tian2020shape}~{\color{gray}[ECCV2020]}                     & \usym{2713}                            & 32.5  & 2.6   & 3.5   & 13.9   \\
SGPA~\cite{chen2021sgpa}~{\color{gray}[ICCV2021]}                          & \usym{2713}                            & 63.6  & 26.2  & 29.2  & 39.5   \\
DualPoseNet~\cite{lin2021dualposenet}~{\color{gray}[ICCV2021]}        & \usym{2717}                            & 70.0  & 17.8  & 22.8  & 36.5   \\
CR-Net~\cite{wang2021category}~{\color{gray}[IROS2021]}               & \usym{2713}                            & 49.5  & 16.1  & 19.2  & 36.4   \\
RePoNet~\cite{ze2022category}~{\color{gray}[NeurlPS2022]}             & \usym{2713}                            & \underline{70.3}  & 29.5  & 34.4  & \underline{42.5}   \\
GPV-Pose~\cite{di2022gpv}~{\color{gray}[CVPR2022]}                    & \usym{2717}                            & 67.8  & 14.1  & 21.5  & 41.1   \\
GPT-CORE~\cite{zou2023gpt}~{\color{gray}[TCSVT2023]}                  & \usym{2713}                            & 66.1  & 29.8  & \underline{35.6}  & 42.3   \\ \midrule
Ours                                                                  & \usym{2713}                            & \textbf{75.1}  & \textbf{31.2}  & \textbf{44.4}  & \textbf{50.9}   \\ \bottomrule [0.8pt]
\end{tabular}

\end{table}
\begin{table}[h]
\scriptsize
\footnotesize
\setlength{\tabcolsep}{1.5pt}
\centering
\renewcommand\arraystretch{1.1}
\caption{\textbf{Pose tracking speed in FPS.} Note that the best and the second best results are highlighted in \textbf{bold} and \underline{underlined}. All speeds are measured on a single NVIDIA RTX A6000 GPU.}
\label{table_FPS}
\vspace{-0.2cm}
\begin{tabular}{c|ccc|cc|c}
\toprule [0.8pt]
Method       & \begin{tabular}[c]{@{}c@{}}NOCS\\~\cite{wang2019normalized}\end{tabular} & \begin{tabular}[c]{@{}c@{}}SPD\\~\cite{tian2020shape}\end{tabular} & \begin{tabular}[c]{@{}c@{}}SGPA\\~\cite{chen2021sgpa}\end{tabular} & \begin{tabular}[c]{@{}c@{}}6-PACK\\~\cite{wang20206}\end{tabular} & \begin{tabular}[c]{@{}c@{}}CAPTRA\\~\cite{weng2021captra}\end{tabular} & Ours  \\ \midrule
Type         & \multicolumn{3}{c|}{Track-free}                                                                                                                                                & \multicolumn{2}{c|}{Track-based}                                                                                                                                                     & Track-based \\ \midrule
NOCS-REAL275 & 5.24        & \underline{15.23}   & 14.12            & 4.03      & 10.35        & \textbf{24.20}  \\
Wild6D       & 5.44        & \underline{14.22}   & 13.58            & 4.98      & 11.23        & \textbf{23.83}  \\
YCB-Video    & 6.39        & \underline{15.74}   & 14.52           & 5.01      & 12.44        & \textbf{23.27}  \\ \bottomrule [0.8pt]
\end{tabular}\vspace{-0.2cm}
\end{table}

\begin{figure}[ht]
    \centering
    \includegraphics[width=0.49\textwidth]{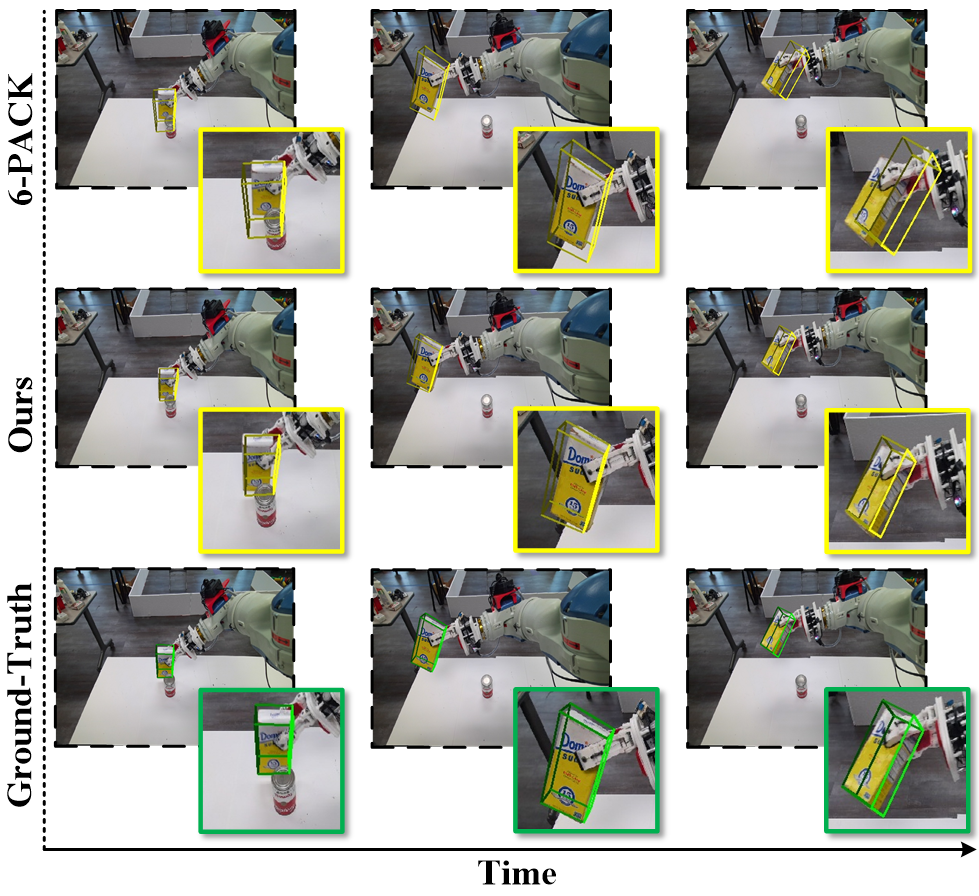} 
    \vspace{-0.6cm}
    \caption{\textbf{Visualization comparison on YCBInEOAT dataset.} We compare our proposed \emph{Robust6DoF} with representative baselines (6-PACK~\cite{wang20206}). Yellow and green represent the results from prediction and ground-truth label, respectively.}\label{FIG_show_ycbineoat}\vspace{-0.2cm}
\end{figure}

\begin{figure}[ht]
    \centering
    \includegraphics[width=0.49\textwidth]{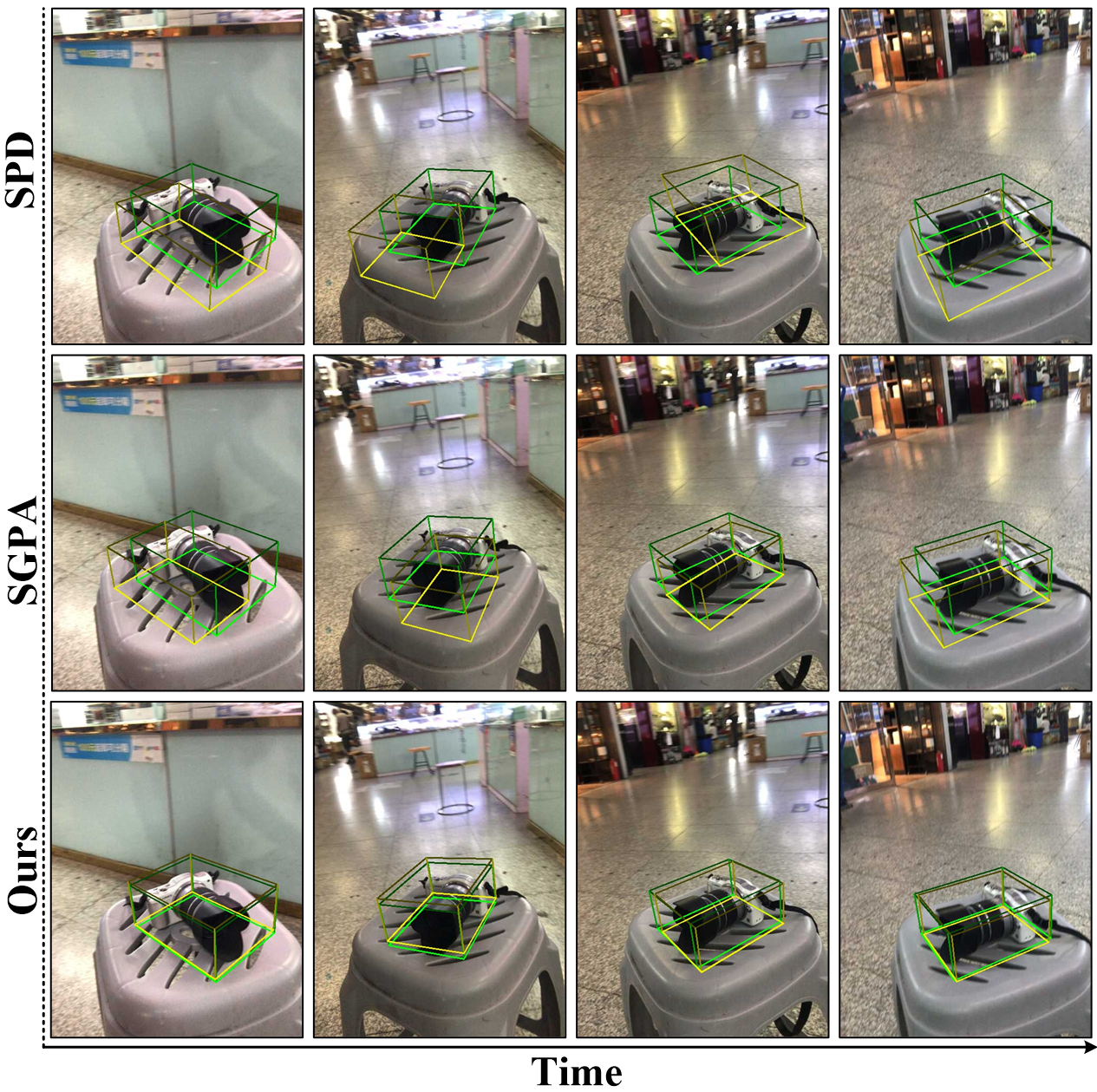} 
    \vspace{-0.6cm}
    \caption{\textbf{Visualization comparison on Wild6D dataset.} We compare our approach with representative baselines (SGPA~\cite{chen2021sgpa} and SPD~\cite{tian2020shape}). Yellow and green represent the results from prediction and ground-truth.}\label{FIG_show_wild6d}\vspace{-0.5cm}
\end{figure}

\subsubsection{Results on the YCB-Video Dataset}
To futher verify the generalization ability of our method regarding the instance-level pose tracking, we verify our model without fine-tuning on the YCB-Video dataset's testing set. We compared our performance with other relevant instance-level detect-based (PoseCNN~\cite{xiang2017posecnn}) and track-based (CatTrack~\cite{yu2023cattrack}) baselines, presenting the average results across all $21$ classes in TABLE~\ref{table_ycb_video_compare}. It can be observed that our method performs well in improving performance for object pose tracking. Specifically, our model without fine-tuning achieved the highest average accuracy of $83.4\%$ and $85.6\%$ in ADD and ADD-S metrics, respectively. Visualization comparison results between our predictions and PoseCNN~\cite{xiang2017posecnn} are also provided in Fig.~\ref{FIG_show_ycb-video}. It also demonstrates that our proposed approach can predict higher-quality pose tracking results for unseen objects. 
\subsubsection{Results on the YCBInEOAT Dataset}
To verify the effectiveness of 6-DoF pose tracking in desktop-fixed robotics manipulation scenarios and to evaluate the performance in situations where objects are moving in front of camera, we compare our \emph{Robust6DoF} on the YCBInEOAT dataset with several available baselines, including 3D model-based methods (RGF~\cite{issac2016depth} and POT~\cite{wuthrich2013probabilistic}) and model-free methods (MaskFusion~\cite{runz2018maskfusion} and TEASER~\cite{yang2020teaser}). Corresponding quantitative and qualitative results are displayed in TABLE~\ref{table_ycbineoat_compare} and Fig.~\ref{FIG_show_ycbineoat}. Overall, \emph{Robust6DoF} achieves the best performance in two average metrics. Specifically, we outperforms TEASER, the latest state-of-the-art baseline, by 27.5\% at ADD and 9.1\% at ADD-S.
As shown in the figure, the pose tracking results by our \emph{Robust6DoF} exhibit a closer value to the ground-truth compared to the 6-PACK~\cite{wang20206} results. These analyses demonstrate that our proposed method not only achieves the better performance for static objects but also facilitates the superior generalizability for dynamic instances captured by a fixed camera.

\subsubsection{Results on the Wild6D Dataset}
To assess the generalization ability of our method in handling overcrowded objects in real-world cluttered scenes, we conduct evaluations on the public Wild6D dataset. We directly test our trained model with some existing works as reported in TABLE~\ref{table_wild6d}. Their pre-trained models, trained on NOCS-REAL275 along with CAMERA75 datasets~\cite{wang2019normalized}, were used for comparison.  It is observed that our proposed achieves 75.1\%, 31.2\%, 44.4\% and 50.9\% on $IoU50$, ${5^ \circ }2cm$, ${5^ \circ }5cm$ and ${10^ \circ }5cm$, respectively, outperforming these available baselines on almost all metrics. This significant improvement shows the superior generalization ability of our approach under the crowded settings in the wild. Additionally, we perform a qualitative comparison of pose tracking by our method and SGPA~\cite{chen2021sgpa} and SPD~\cite{tian2020shape} on the Wild6D testing set. The results are displayed in Fig.~\ref{FIG_show_wild6d}. It can be concluded that we can exhibit a closer match to the ground-truth compared to existing single estimation method SGPA and SPD. These analysis and results showcase the potential of our \emph{Robust6DoF}.

\subsubsection{Pose Tracking Speed in FPS}
Beyond the comparison of performance with state-of-the-arts, we futher verify the tracking speed (FPS) among five typical baselines: NOCS~\cite{wang2019normalized}, SPD~\cite{tian2020shape}, SGPA~\cite{chen2021sgpa}, 6-PACK~\cite{wang20206} and CAPTRA~\cite{weng2021captra}. As summarized in TABLE~\ref{table_FPS}, all methods are tested on the same device using their officially released code or checkpoint to ensure a fair evaluation. From TABLE~\ref{table_FPS}, it is evident that our method achieves an average speed of 24.2 FPS on the NOCS-REAL275 dataset, 23.8 FPS on the Wild6D dataset and 23.3 FPS on the YCB-Video dataset, respectively. It is clear that our method outperforms these existing track-based and track-free approaches.

\begin{figure}[t]
    \begin{minipage}[t]{\linewidth}
        \centering
        \includegraphics[width=0.85\textwidth]{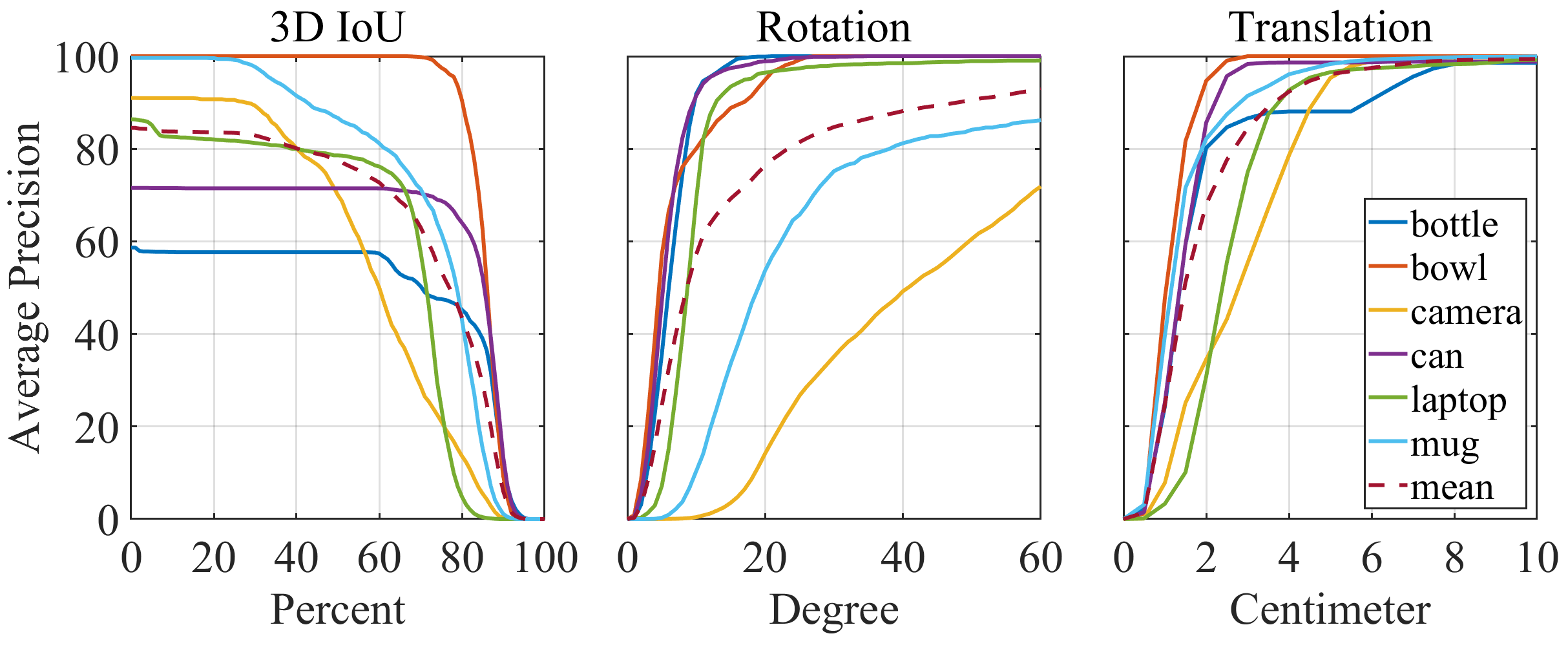}
        \centerline{(a) SPD~\cite{tian2020shape} results.}
    \end{minipage}%
    \vspace{0.2cm}
    \quad
    \begin{minipage}[t]{\linewidth}
        \centering
        \includegraphics[width=0.85\textwidth]{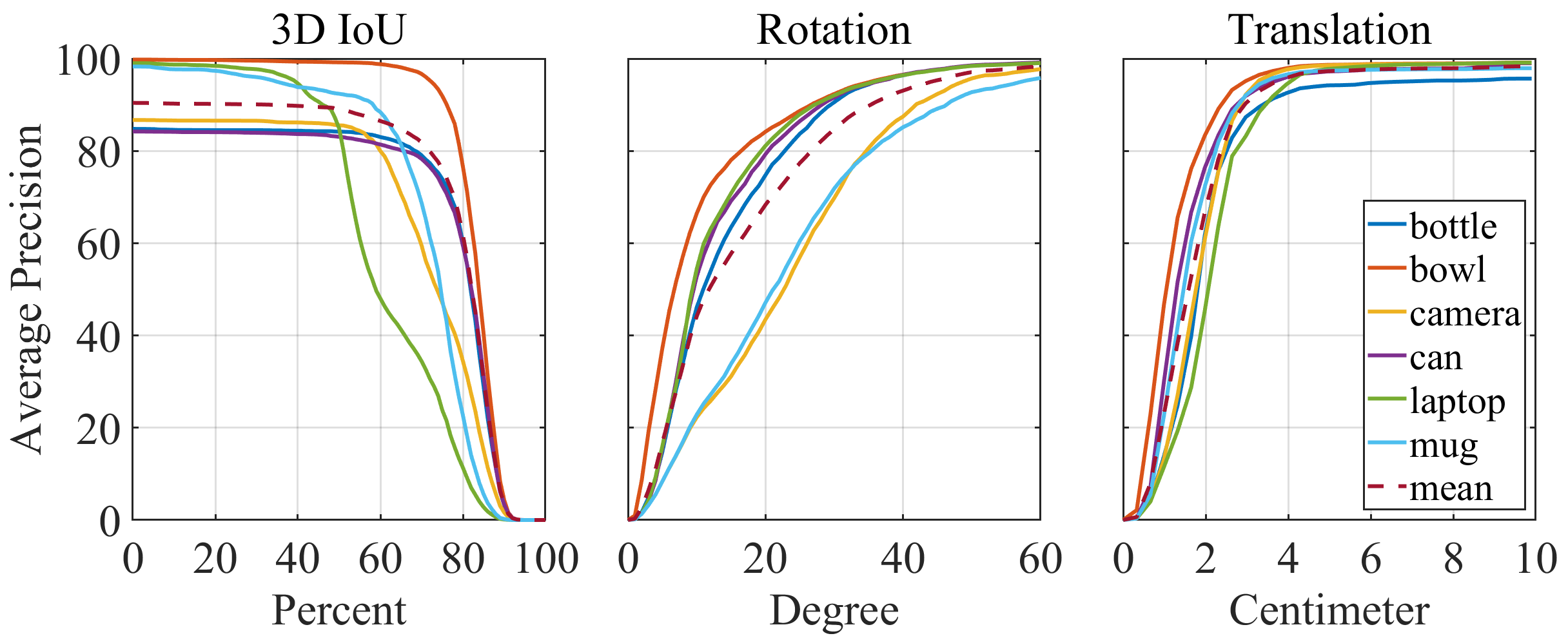}
        \centerline{(b) Our results.}
    \end{minipage}    
    \caption{\textbf{Comparison of mAP on public NOCS-REAL275 dataset.} Mean Average Percision (mAP) of our \emph{Robust6DoF} and representative baseline SPD~\cite{tian2020shape} for various 3D IoU, rotation and translation errors. }
    \label{fig-map}
    \vspace{-0.2cm}
\end{figure}

\begin{figure}[t]

\centering
\includegraphics[width=0.475\textwidth]{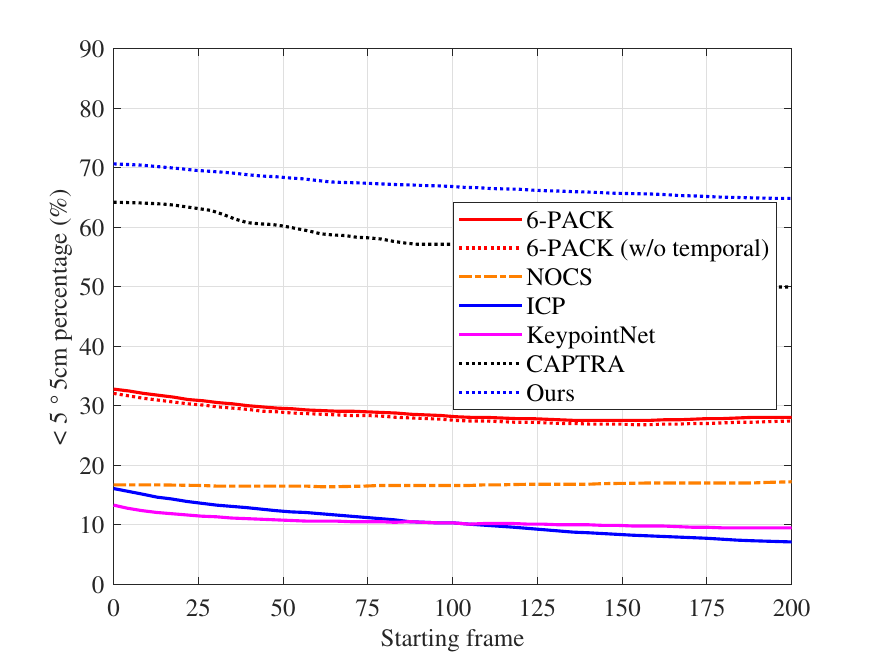}\vspace{-0.2cm}
\caption{\textbf{Robustness evaluation of frame drops over time.} Each point on the x-axis represents the number of consecutive frames lost between the initial frame and the second frame, and each point on the curve represents the mean success rate (${5^ \circ }5cm$ percentage) on the interval of the sequence without lost frame on the x-axis. }
\label{fig-sta}
\vspace{-0.2cm}
\end{figure}

\begin{figure}[t]
\centering

\includegraphics[width=0.49\textwidth]{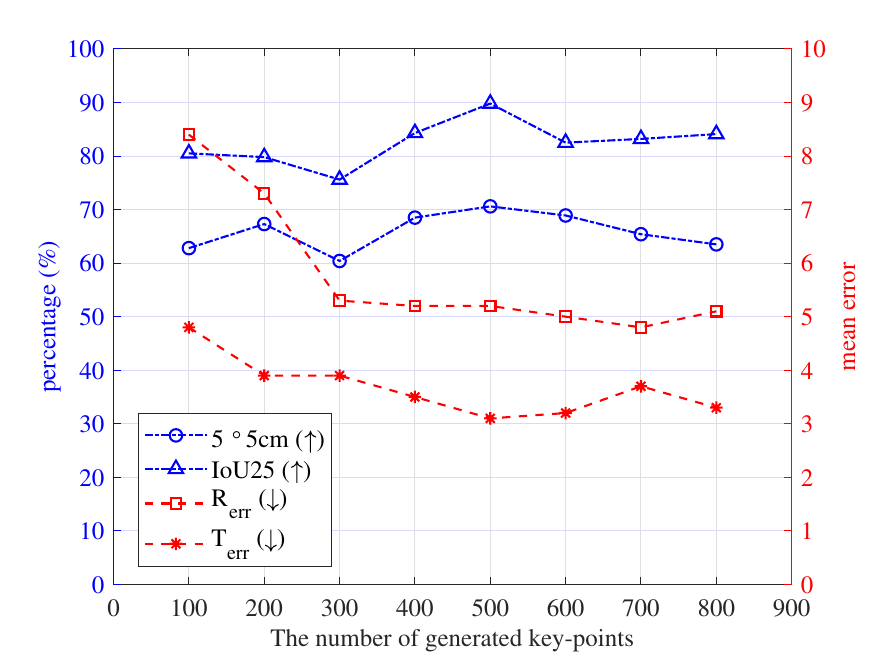}\vspace{-0.2cm}
\caption{\textbf{Sensitivity evaluation of different number of keypoints.} x-axis indicates the number of generated keypoints. The left y-axis (blue) represents the percentage of metrics ${5^ \circ }5cm$ and $IoU25$, and the right y-axis (red) represents the means of the error in ${R_\mathit{err}}$ and ${T_\mathit{err}}$. The results are averaged over all six categories.}
\label{fig-sen}
\vspace{-0.2cm}
\end{figure}

\begin{figure}[t]

\centering
\includegraphics[width=0.49\textwidth]{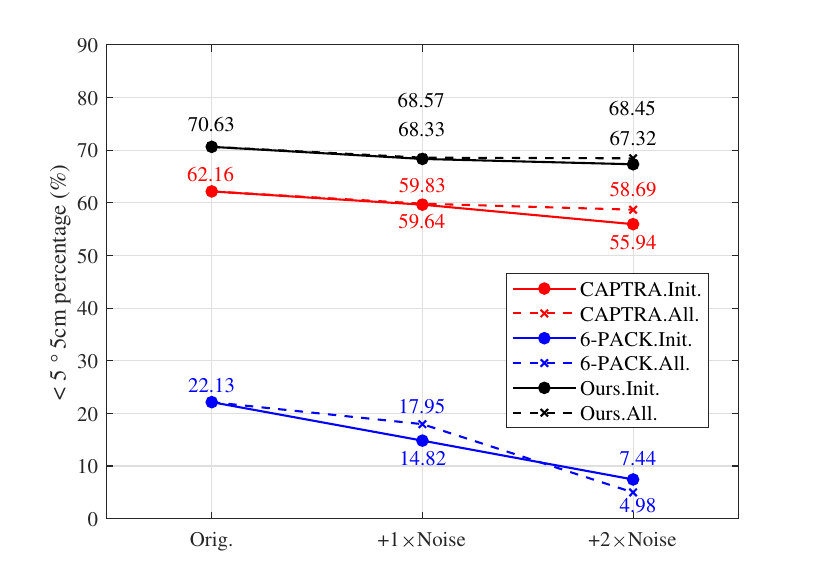}\vspace{-0.2cm}
\caption{\textbf{Tracking robustness evaluation against noisy pose inputs.} The $+n\times Noise$ on the x-axis represents adding $n$ time noise during training time. $Orig.$ denotes the original setting, $Init.$ denotes the only for inital pose and $All.$ denotes every frame during training.  The results of the comparison methods are summarized from \cite{weng2021captra}.}
\label{fig-rob}
\vspace{-0.2cm}
\end{figure}
\subsection{Additional Analyses}
To assess the pose tracking robustness of the proposed \emph{Robust6DoF}, we also conduct several extra experiments on the NOCS-REAL275 dataset, the detailed results are displayed in Fig.~\ref{fig-map} to Fig.~\ref{fig-rob}.

\subsubsection{Comparison of Mean Average Precision (mAP)}
To futher analyze the performance of our method for various instances with the same category, we also present detailed per-category results for 3D IoU, rotation accuracy and translation precision on the NOCS-REAL275 dataset. Meanwhile, to support our claim regarding the generalization robustness of our proposed \emph{Robust6DoF} to the intra-class shape variations, we conduct a quantitative comparison with the related track-free method, SPD~\cite{tian2020shape}. It is evident from the visualization in Fig.~\ref{fig-map} that we outperforms SPD~\cite{tian2020shape} in mean accuracy for almost all thresholds, especially
in both two evaluation metrics: 3D IoU and translation estimation. 

\subsubsection{Stability to Dropped Frames}
Here, we examine how the frame dropping affects tracking performance. We drop the next $N$ frames after the first frame and use the mean performance of ${<5^ \circ }5cm$ to evaluate different baselines, as depicted in the Fig.~\ref{fig-sta}. The fewer frames dropped after the first frame, the easier it is to track. Note that the performance of almost all methods decreases as the number of dropped frames increases, except for NOCS because it is a track-free method, that is not influenced by dropped frames.
Compared to other track-based baselines, our method decreases only $3.2\%$ when dropping 75 frames, while the state-of-the-art CAPTRA~\cite{weng2021captra} reduces by $5\%$. Meanwhile, the performance of our method is $10\%$ higher than CAPTRA throughout the whole process, and remains steady at about 100 frames.

\subsubsection{Tracking Comparison to Pose Noise}\label{pose noise sec}
This experiment validates the performance of our method with the noisy pose inputs. Randomly sampled pose noise is added during training time. As shown in Fig.~\ref{fig-rob}, we directly test our method under the following settings:
(i)~increasing the pose noise by $1$ or $2$ times, denoted as $ + n \times Noise(n = 1,2)$; 
(ii)~adding the pose noise in the initial pose, denoted as $Init.$, 
and (iii)~adding the pose noise to pose prediction of every previous frame, denoted as $All.$. 
We compare our method with state-of-the-art methods, 6-PACK~\cite{wang20206} and CAPTRA~\cite{weng2021captra} under the same settings, and show the results under the metric ${5^ \circ }5cm$ in Fig.~\ref{fig-rob}. It can be seen that our method is more robust to pose noise than other baseline methods.

\subsubsection{Sensitivity to the Number of Generated Keypoints}
We also evaluate the sensitivity of our proposed method with a different number of generated keypoints, as shown in Fig.~\ref{fig-sen}. We chose eight different sets of generated keypoint numbers, ranging from 100 to 800. Our \emph{Robust6DoF} with about $n = 500$ (we set $n = 512$ for experiments), achieves optimal performance, and our model doesn't seem to be very sensitive to this parameter.

\subsection{Ablation Studies}

\begin{table}[t]
\scriptsize
\footnotesize
\setlength{\tabcolsep}{1.5pt}
\centering
\renewcommand\arraystretch{1.0}
\caption{\textbf{Ablation experiments on different components of our \emph{Robust6DoF}.} Note that "WAS", "GDF", "STE", "SSF" and "AD" represent the weight-shared attention, global dense fusion, spatial-temporal filtering encoder, shape-similarity filtering and augmentation decoder,respectively.}
\vspace{-0.2cm}
\label{ablation-components}
\begin{tabular}{c|ccccc|cccc|cc}
\toprule [0.8pt]
\multirow{2}{*}{\#} & \multirow{2}{*}{WAS} & \multirow{2}{*}{GDF} & \multirow{2}{*}{STE} & \multirow{2}{*}{SSF} & \multirow{2}{*}{AD} & \multicolumn{4}{c|}{NOCS-REAL275} & \multicolumn{2}{c}{Wild6D}                                              \\
                    &                      &                     &                      &                      &                     & $IoU25$ & ${5^ \circ }2cm$ & \multicolumn{1}{l}{${R_{err}}$} & \multicolumn{1}{l|}{${T_{err}}$} & $IoU50$ & ${5^ \circ }2cm$ \\ \midrule 
M0                  & \usym{2717} & \usym{2717} & \usym{2717} & \usym{2717} & \usym{2717} & 63.5  & 30.1  & 19.6 & 14.3 & 57.6 & 20.7                        \\
M1                  & \usym{2713}                    & \usym{2717}                   & \usym{2717}                    & \usym{2717}                    & \usym{2717}                   & 70.0  & 33.2  & 14.3                     & 9.9                      & 62.0          & 22.7                                  \\
M2                  & \usym{2713}                    & \usym{2713}                   & \usym{2717}                    & \usym{2717}                    & \usym{2717}                   & 71.2  & 33.5  & 14.8                     & 10.6                      & 62.3          & 23.0                                  \\
M3                  & \usym{2713}                    & \usym{2713}                   & \usym{2713}                    & \usym{2717}                    & \usym{2717}                   & 80.8  & 50.4  & 8.9                     & 7 .3                      & 60.0          & 25.8                                  \\
M4                  & \usym{2713}                    & \usym{2713}                   & \usym{2713}                    & \usym{2713}                    & \usym{2717}                   & \textbf{90.2}  & 55.4  & 5.5                     & 3.7                      & 74.8          & 30.0                                  \\ \midrule 
M5                  & \usym{2713}                    & \usym{2713}                   & \usym{2713}                    & \usym{2713}                    & \usym{2713}                   & 89.8  & \textbf{57.1}  & \textbf{5.2}                     & \textbf{3.0}                      & \textbf{75.1}          & \textbf{31.2}                                  \\ \bottomrule [0.8pt]
\end{tabular}
\end{table}

\begin{table}[t]
\scriptsize
\footnotesize
\setlength{\tabcolsep}{0.5pt}
\centering
\renewcommand\arraystretch{1.0}
\caption{\textbf{Ablation study on keypoints generation manner of our \emph{Robust6DoF}.} "unsupervised KP" means we use the unsupervised keypoints generation method introduced in 6-PACK~\cite{wang20206}.}
\vspace{-0.2cm}
\label{ablation-keypoints}
\begin{tabular}{c|c|cccc}
\toprule [0.8pt]
\multirow{2}{*}{Dataset}      & \multirow{2}{*}{Settings} & \multicolumn{4}{c}{Evaluation Metrics} \\
                              &                           & $IoU50$   & ${5^ \circ }2cm$   & ${5^ \circ }5cm$   & ${10^ \circ }5cm$   \\ \midrule
\multirow{4}{*}{NOCS-REAL275} & Ours w/o prior guidance   & 58.6    & 33.3    & 35.9    & 62.2     \\
                              & Ours w/o finer matching   & 70.8    & 50.4    & 62.5    & 66.7     \\
                              & Ours + unsupervised KP    & 80.2    & 55.4    & 63.0    & 83.0     \\
                              & Ours                      & \textbf{87.0}    & \textbf{57.1}    & \textbf{70.6}    & \textbf{84.5}     \\ \midrule
\multirow{4}{*}{Wild6D}       & Ours w/o prior guidance   & 65.2    & 26.1    & 29.6    & 35.0     \\
                              & Ours w/o finer matching   & 72.1    & 29.4    & 43.2    & 44.8     \\
                              & Ours + unsupervised KP    & 73.5    & 30.4    & 40.1    & 48.9     \\
                              & Ours                      & \textbf{75.1}    & \textbf{31.2}    & \textbf{44.4}    & \textbf{50.9}     \\ \bottomrule [0.8pt]
\end{tabular}
\end{table}

\begin{table}[t]
\scriptsize
\footnotesize
\setlength{\tabcolsep}{2.2pt}
\centering
\renewcommand\arraystretch{1.0}
\caption{\textbf{Ablation study on loss functions of our \emph{Robust6DoF}.} "${L_{base}}$" contains both rotation loss "${L_{rot}}$" and translation loss "${L_{tra}}$".}
\vspace{-0.2cm}
\label{ablation-loss}
\begin{tabular}{c|cccc|cccc|cc}
\toprule [0.8pt]
\multirow{2}{*}{\#} & \multirow{2}{*}{${L_{base}}$} & \multirow{2}{*}{${L_{aux}}$} & \multirow{2}{*}{${L_{mvc}}$} & \multirow{2}{*}{${L_{c}}$} & \multicolumn{4}{c|}{NOCS-REAL275} & \multicolumn{2}{c}{Wild6D} \\
                    &                        &                       &                       &                     & $IoU25$   & ${5^ \circ }2cm$   & ${R_{err}}$  & ${T_{err}}$  & $IoU50$        & ${5^ \circ }2cm$       \\ \midrule
\ding{172}                   & \usym{2713}                      & \usym{2717}                     & \usym{2717}                     & \usym{2717}                               & 82.2    & 48.7    & 10.2  & 9.3  & 68.6         & 25.4        \\
\ding{173}                   & \usym{2713}                      & \usym{2713}                     & \usym{2717}                     & \usym{2717}                               & 85.2    & 51.3    & 8.0  & 7.5  & 71.1         & 28.9        \\
\ding{174}                   & \usym{2713}                      & \usym{2713}                     & \usym{2713}                     & \usym{2717}                               & 87.0    & 54.2    & 6.6  & 5.3  & 72.1         & 30.0        \\ \midrule
\ding{175}                   & \usym{2713}                      & \usym{2713}                     & \usym{2713}                     & \usym{2713}                               & \textbf{89.8}    & \textbf{57.1}    & \textbf{5.2}  & \textbf{3.0}  & \textbf{75.1}         & \textbf{31.2}        \\ \bottomrule [0.8pt]
\end{tabular}
\end{table}

\begin{table}[t]

\scriptsize
\footnotesize
\setlength{\tabcolsep}{3.0pt}
\centering
\renewcommand\arraystretch{1.0}
\caption{\textbf{Ablation study on robustness to pose errors of our \emph{Robust6DoF}.} "Init.~$\times n$" and "All.~$\times n$" means adding $n$ $(n = 1,2)$ times train-time errors in initial pose and adding $n$ times pose errors to all frames.}
\vspace{-0.2cm}
\label{ablation-noise}
\begin{tabular}{c|c|c|cc|cc}
\toprule [0.8pt]
Dataset                                                                 & Metric & Orig. & Init.~$\times1$ & Init.~$\times2$ & All.~$\times1$ & All.~$\times2$ \\ \midrule
\multirow{4}{*}{\begin{tabular}[c]{@{}c@{}}NOCS-REAL275\end{tabular}}   & $IoU25$           & \textbf{89.9}  & 88.8    & 87.1    & 88.2   & 87.9   \\
                                                                        & ${5^ \circ }5cm$  & \textbf{70.6}  & 68.3    & 67.3    & 68.6   & 68.5   \\
                                                                        & ${R_{err}}$       & \textbf{5.2}   & 5.57    & 5.64    & 5.61   & 5.79   \\
                                                                        & ${T_{err}}$       & \textbf{3.0}   & 3.94    & 3.94    & 4.03   & 4.98   \\ \midrule
\multirow{2}{*}{Wild6D}                                                 & $IoU50$           & \textbf{75.1}  & 73.4    & 72.7    & 74.0   & 72.8   \\
                                                                        & ${5^ \circ }5cm$  & \textbf{44.4}  & 43.0    & 42.1    & 42.8   & 42.6   \\ \bottomrule [0.8pt]
\end{tabular}
\end{table}

\subsubsection{Effectiveness Evaluation of Different Components}

To evaluate the effectiveness of the dividual components in our \emph{Robust6DoF}, we conducted ablation studies and presented the results on public datasets \emph{i.e.,} NOCS-REAL275 and Wild6D in TABLE~\ref{ablation-components}. 
We start with a base model and incrementally add each proposed component to this baseline. This base model, denoted as "M0", is built using the classical Scaled Dot-Product Multi-Head Attention and the Transformer in~\cite{vaswani2017attention}, along with the proposed keypoints generation and match module, and the training strategy is consistent with \emph{Robust6DoF}.
First, the results of "M1" and "M2" in TABLE~\ref{ablation-components} show that incorporating the WSA layer into the base model resulted in a significant performance improvement, demonstrating the effectiveness of proposed 2D-3D Dense Fusion module. Secondly, by comparing "M0", "M3" and "M5", we can observe that the proposed Spatial-Temporal Filtering Encoder can provide efficient dynamic enhancement to capture the temporal information and improve the inference ability.
Our third experiment aims to verify the effectiveness of the proposed Augmentation Decoder with the shape-similarity filtering, as shown in "M4" and "M5". Without the "SSF" and "AD" block, the pose tracking performance would be severely weakened.
Our complete model "M5" outperforms all other variants in all comparison experiments.  

\subsubsection{Comparison of Different Keypoints Generations}
We also compare our proposed Prior-Guided Keypoints Generation and Match module with its three different manners: (i) keypoint generation without prior guidance, (ii) our method using only the initial matching, and (iii) unsupervised keypoints generation in 6-PACK~\cite{wang20206}. As presented in TABLE~\ref{ablation-keypoints}, the results in both (i) and (ii) manners simultaneously perform the worst, while our proposed approach has the best performance. It can also be seen that unsupervised manner (iii) is slightly better, but its tracking robustness is significantly worse due to the lack of shape prior's supervision. These experiment results indicate that our proposed manner is more effective in capturing the changes in category relationships among different instances, making it more suitable for category-level pose tracking.

\subsubsection{Impact of Different Loss Configurations}
In TABLE~\ref{ablation-loss}, we compare the generalization capability under different loss combinations during the training stage. We start with the base losses, including ${L_{rot}}$ and ${L_{tra}}$, and incrementally add other losses in order. The experimental results in \#\ding{172} and \#\ding{173} demonstrate that the prior-guided auxiliary module is very important for keypoints generation. The results in \#\ding{173} and \#\ding{174} indicate that the supervision of keypoint's consistency is also critical to improve performance. 
Furthermore, we also explore the impact of the proposed key-points refine matching block with the loss ${L_{c}}$. The results in \#\ding{174} and \#\ding{175} show its crucial role in capturing the more critical key-points and catching the the structural changes between observable points and prior-points. Finally, our model \#\ding{175} achieves the best performance under all loss supervisions. 

\subsubsection{Robustness of Additional Pose Noises}
TABLE~\ref{ablation-noise} shows the detailed ablation experiments of our \emph{Robust6DoF} with respect to the added pose noise on NOCS-REAL275 and Wild6D datasets. Following the same settings in subsection~\ref{pose noise sec}, we further verify our \emph{Robust6DoF} perfromance to examine the tracking robustness to extra pose noises, where we add one or two times pose noises into initial pose or each pose prediction in every frame. As shown in TABLE~\ref{ablation-noise}, the tracking performance of our model is steadily weakening without a particularly severe decrease, which further demonstrates the robustness of our proposed \emph{Robust6DoF}. 
The visualization comparison with 6-PACK~\cite{wang20206} and CAPTRA~\cite{weng2021captra} is displayed in Fig.~\ref{fig-rob}.

\subsection{Real-World Experiments on an Aerial Robot}
\begin{figure}[ht]
    \centering
    \includegraphics[width=0.49\textwidth]{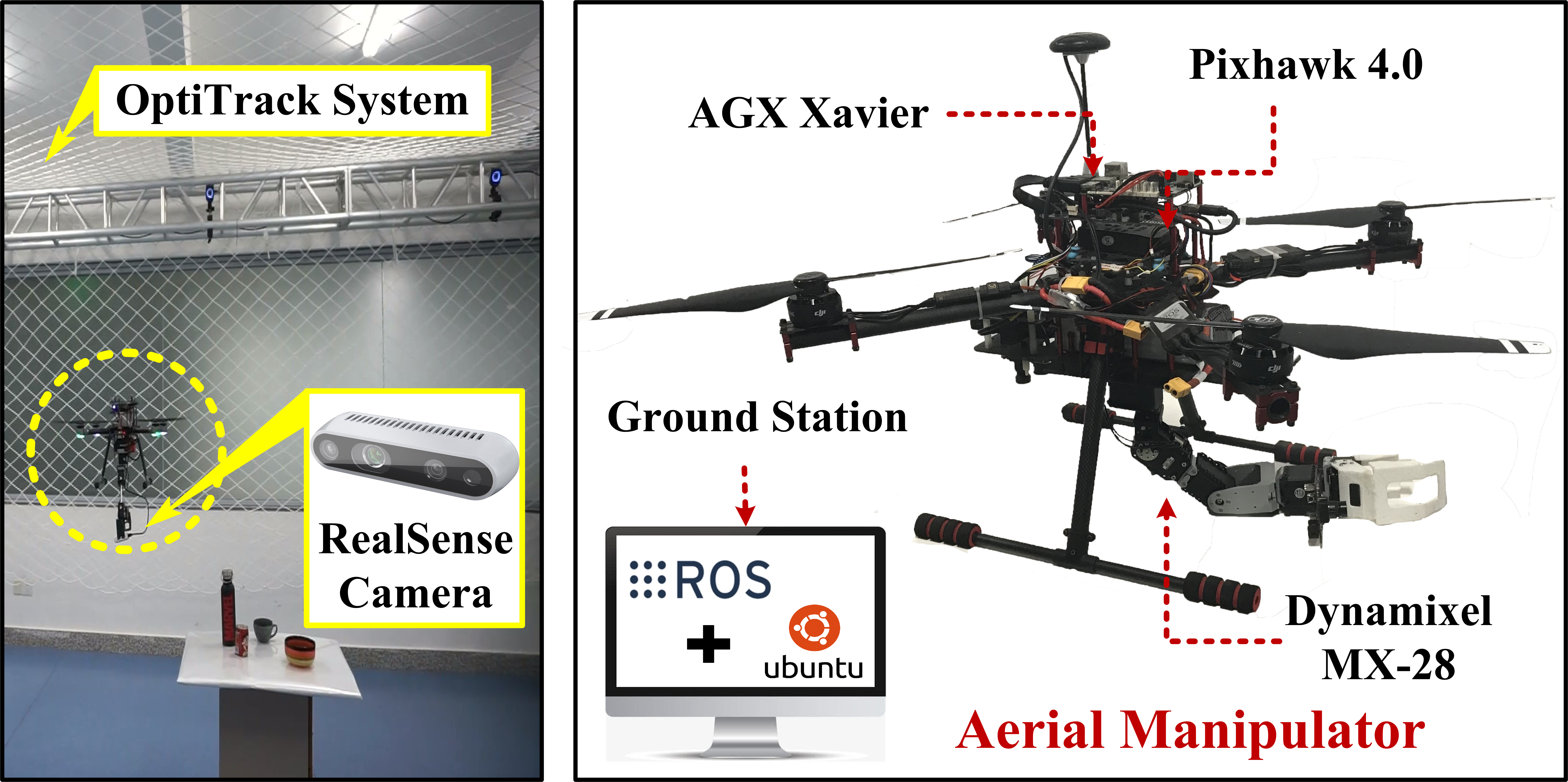} 
    \caption{\textbf{Real-world experimental setup.} All ground-truth information comes from measurements of every electronic unit.}\label{FIG_real_show}    
\end{figure}
In addition to the quantitative experiments for the proposed 6-DoF pose tracker, we further test the complete algorithm in a real-world experiment using the aerial robot developed in our Robotic Laboratory. As shown in Fig.~\ref{FIG_real_show}, the entire experiment platform includes an OptiTrack indoor motion capture system and an Aerial Manipulator, which is mainly composed of a quadrotor, a 4-DoF robotic manipulator and a downward-looking RGB-D camera (RealSense D435i) to capture the real-time image data. The OptiTrack motion capture system communicates with ground station through WiFi to record the ground-truth position information of the quadrotor with respect to the global coordinate frame. A custom-made electronics flight controllor board (Pixhawk 4.0 with IMUs) provides the ground-truth angle and veocity information of quadrotor, and an onboard computer (Jetson AGX Xavier) runs the closed-loop control the whole system at 20 HZ. This onboard computer also records the ground-truth angle veocity information generated from the robotic manipulator (four Dynamixel MX-28). The proposed algorithm is implemented under the Robot Operating System with Ubuntu 20.04.

\begin{figure*}[ht]
    \centering
    \includegraphics[width=\textwidth]{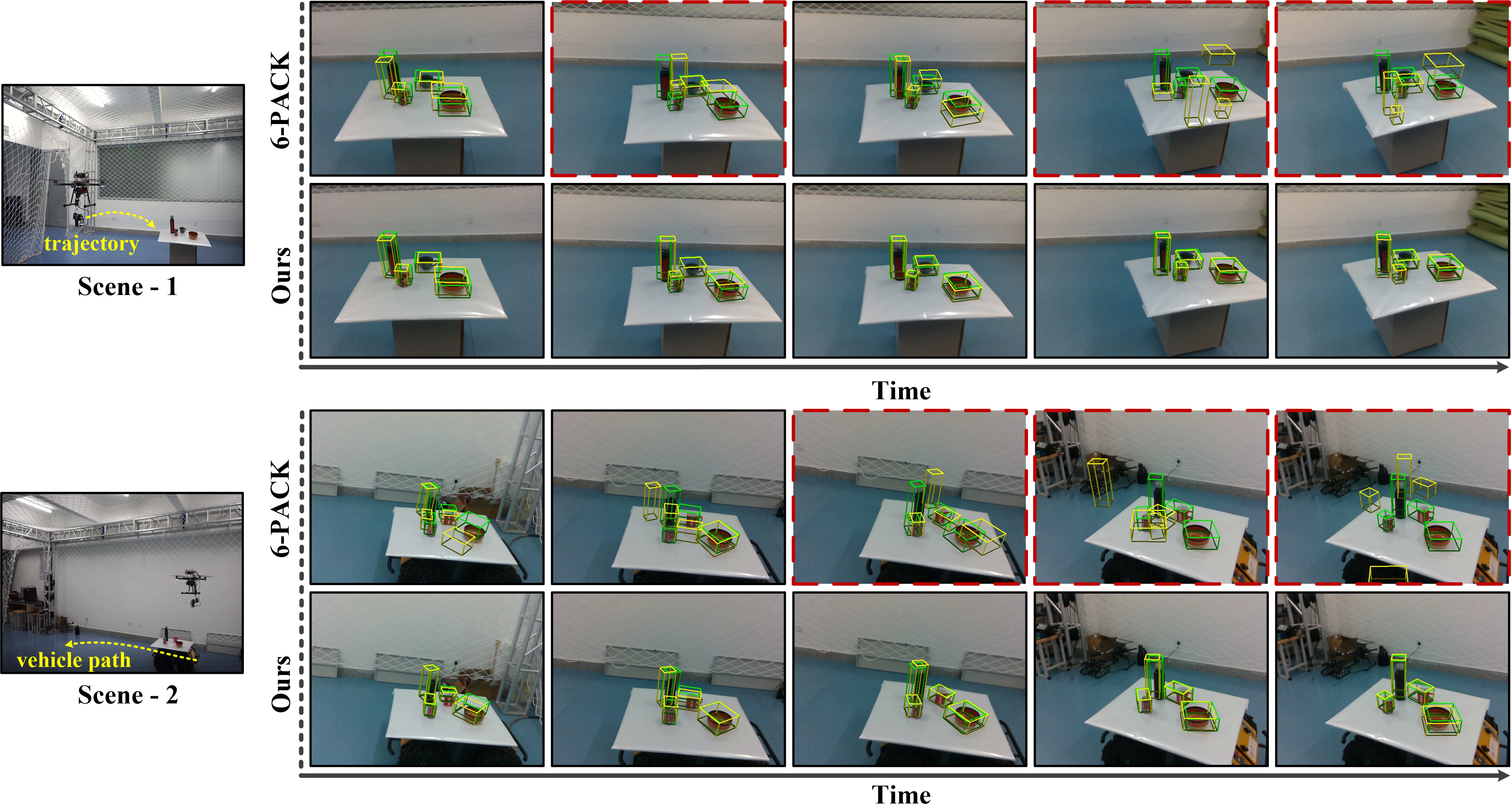} 
     
    \caption{\textbf{Real visualization comparison.} We compare with 6-PACK~\cite{wang20206}. Two scenes are considered: 1) table-top fixed objects (upper part); 2) moving objects (bottom part). These results are estimated offline using the recorded real data flow. Yellow and green represent the results from estimations and annotations labeled manually, respectively.}
    \label{FIG_real_track}
\end{figure*}

\begin{figure}[H]

    \centering
    
    \subfigure{
        \centering
        \includegraphics[width=0.48\textwidth]{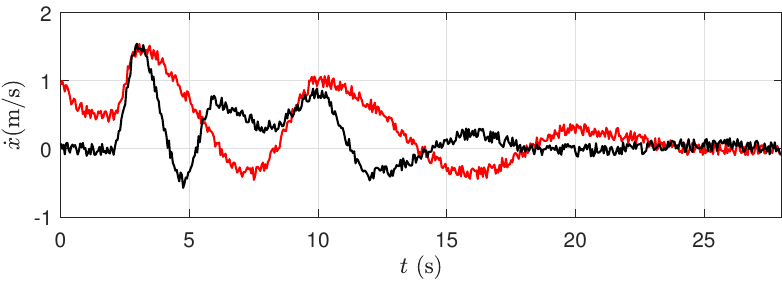}
        \vspace{-0.6cm} 
    }
    \vspace{-0.9cm}
    \qquad
    
    \subfigure{
        \centering
        \includegraphics[width=0.48\textwidth]{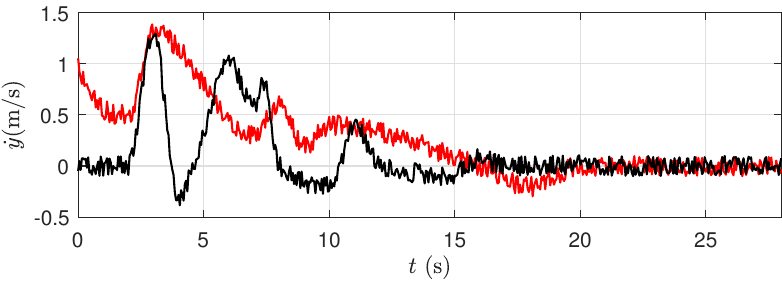}
        \vspace{-0.6cm}     
    }
    \vspace{-0.9cm}
    \qquad
    
    \subfigure{
        \centering
        \includegraphics[width=0.48\textwidth]{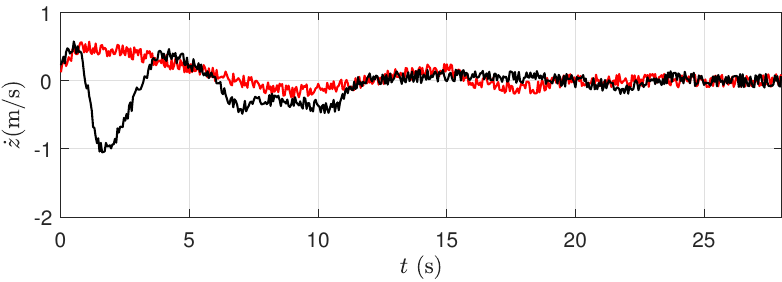}
        \vspace{-0.6cm}
    }
    \vspace{-0.9cm}    
    \qquad
    
    \subfigure{
        \centering
        \includegraphics[width=0.48\textwidth]{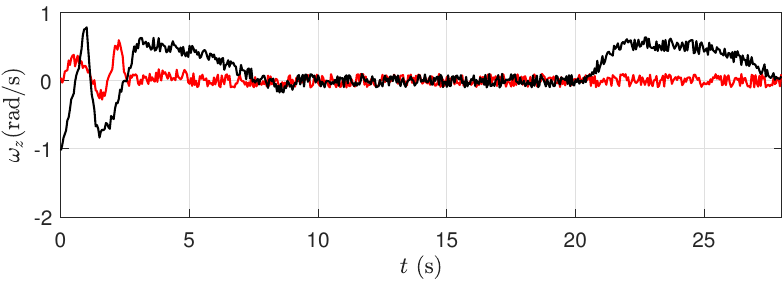}      
    }
    \vspace{-0.8cm} 
    \caption{\textbf{Experiment results of quadrotor velocity vectors ($v$).} The red curve represents the actual outputs generated from our \emph{PAD-Servo}. The black curve represents the corresponding ground-truth historical state measured from devices, where ${v_x},{v_y},{v_z}$ are from OptiTrack and ${\omega _z}$ is from Pixhawk 4.}
    \label{fig_quadrotor}
 \end{figure}

\begin{figure}[H]
    \centering
    
    \subfigure{
        \centering
        \includegraphics[width=0.48\textwidth]{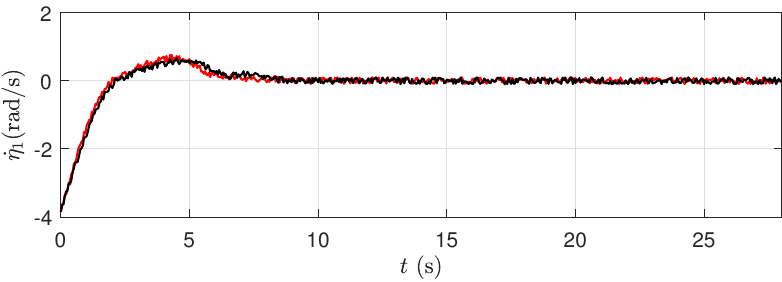}
        \vspace{-0.6cm} 
    }
    \vspace{-0.9cm}
    \qquad
    
    \subfigure{
        \centering
        \includegraphics[width=0.48\textwidth]{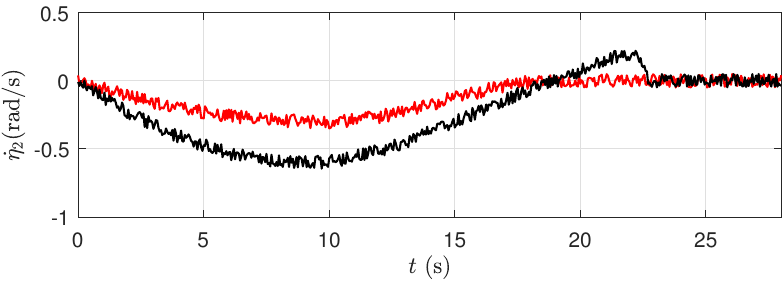}
        \vspace{-0.6cm}     
    }
    \vspace{-0.9cm}
    \qquad
    
    \subfigure{
        \centering
        \includegraphics[width=0.48\textwidth]{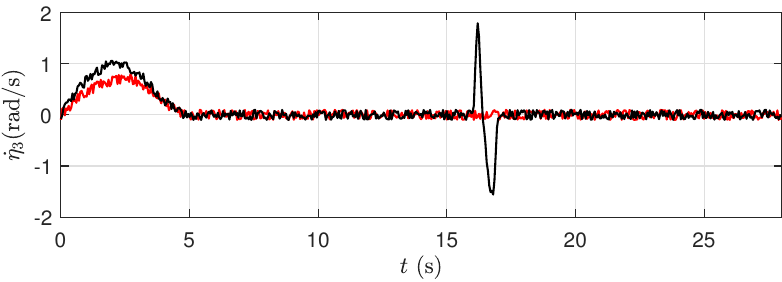}
        \vspace{-0.6cm}
    }
    \vspace{-0.9cm}    
    \qquad
    
    \subfigure{
        \centering
        \includegraphics[width=0.48\textwidth]{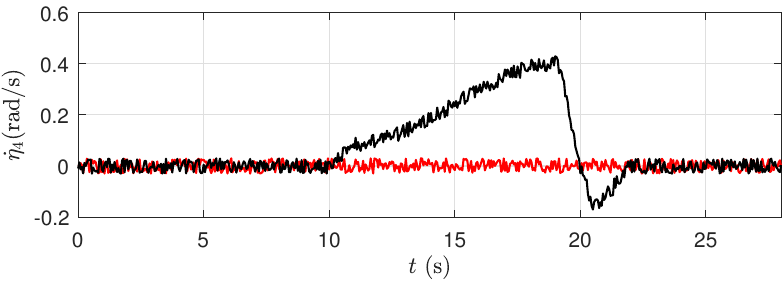}      
    }
    \vspace{-0.8cm} 
    \caption{\textbf{Experiment results of onboard manipulator angle velocity vectors ($\dot \eta $).} The red curve represents the actual outputs generated from \emph{PAD-Servo}. The black curve represents the corresponding ground-truth historical state measured from devices, where ${\dot \eta _1},{\dot \eta _2},{\dot \eta _3},{\dot \eta _4}$ are all from four Dynamixel MX-28 motors.}
    \label{fig_manipulator}
 \end{figure}

We consider two different aerial robotic scenarios, as displayed in Fig.~\ref{FIG_real_track}. The first case involves the aerial manipulator autonomously guiding itself to the neighborhood of fixed objects on a tabletop. The second case entails the aerial manipulator actively following moving objects placed on a ground vehicle. We recorded the original RGB-D data flow online during the experiment and use a offline 3D labeling tool to obtain the
required pose annotations. We compare our method with the representative tracking baseline, 6-PACK~\cite{wang20206}, as shown in the right of Fig.~\ref{FIG_real_track}. 
Druing the begining time, the 6-PACK can detect and estimate each object's pose, but it gradually loses track when the camera's view changes drastically (as depicted in red dotted box). The detailed video will be presented in the project page. In contrast, our \emph{Robust6DoF} achieves robust and effective tracking results. These situation occurs in both two scene cases.
It qualitatively demonstrates that our proposed \emph{Robust6DoF} robust performance in real-world aerial scenarios. We also recorded the action signals output during the process of experiment in first case.
The time evolution of the linear velocity of the quadrotor, mentioned in Eq.~(\ref{vehicle sign}), $\upsilon  = ({v_x},{v_y},{v_z},{\omega _z})^T$ and angle velocity of onboard robotic manipulator, referred in Eq.~(\ref{joint-signs}), $\dot \eta  = {({\dot \eta _1},{\dot \eta _2},{\dot \eta _3},{\dot \eta _4})}$, during the visual guidance process, is shown in Fig.~\ref{fig_quadrotor} and Fig.~\ref{fig_manipulator}, respectively. It can be observed that the quadrotor and manipulator successfully track the reference velocities using the real-time pose estimated by our \emph{Robust6DoF}. The velocity errors converge to a neighborhood around zero without surprise when the whole experimental process comes to an end at $28 s$, where the current 6-DoF object's pose is infinitely close to the desired setting. It converges fast and successfully tracks all reference velocities. All the results show good stability of our \emph{PAD-Servo} scheme and the well real-world performance of our \emph{Robust6DoF} for guiding in aerial robotics manipulation.
\vspace{-0.2cm}
\section{Discussions and Future Works}\label{sec:conclusion}
In this paper, our focus is on an actual robotics task \emph{i.e.,} aerial vision guidance for aerial robotics manipulation. We first proposed a robust category-level 6-DoF pose tracker called \emph{Robust6DoF}, which adopts a three-stage pipeline to achieve aerial object's pose tracking by leveraging the shape prior-guided keypoints alignment. Futhermore, we introduce a pose-aware discrete servo policy for aerial manipulator termed \emph{PAD-Servo}, designed to effectively handle the challenges of real-time dynamic vision guidance task for aerial manipulator. Extensive experiments conducted on four public datasets demonstrate the effectiveness of our proposed \emph{Robust6DoF}. Real-world experiments conducted on our built aerial robotics platform also verify the practical significance of our method, including both the proposed \emph{Robust6DoF} and \emph{PAD-Servo}.

Althought our method has achieved effective and practical real-world performance, there are still many unresolved challenges and limitations in this robotic vision field, such as "how to deal with the sudden appearance and disappearance of objects in the field of onboard camera's view?" and "how to use language, audio, and other multi-modal information to achieve smarter and more autonomous unexplored tasks?", and so on. These are worthy of being explored in our following works.
In our future work, we aim to establish a new dataset to provide researcher with a valueable dataset resource for validating 6-DoF pose tracking in aerial situations. Meanwhile, the challenge of aerial visual pose tracking under the setting of fast view changes remains an open problem. We believe that our work will contribute to the development and further advancement of the aerial robotic vision field. 

\newpage
\newpage

\ifCLASSOPTIONcompsoc
  \section*{Acknowledgments}
\else
  \section*{Acknowledgment}
\fi

This work was supported by the National Key Research and Development Program of China under Grant No.2022YFB3903800, Grant No.2021ZD0114503 and Grant No.2021YFB1714700.
Jingtao Sun is supported by the China Scholarship Council under Grant No.202206130072. 

\ifCLASSOPTIONcaptionsoff
  \newpage
\fi

\bibliographystyle{Bibliography/IEEEtran}
\bibliography{Bibliography/IEEEabrv,bare_jrnl_compsoc} 
\vspace{-0.4cm}
\begin{IEEEbiography}[{\includegraphics[width=1in,height=1.25in,clip,keepaspectratio]{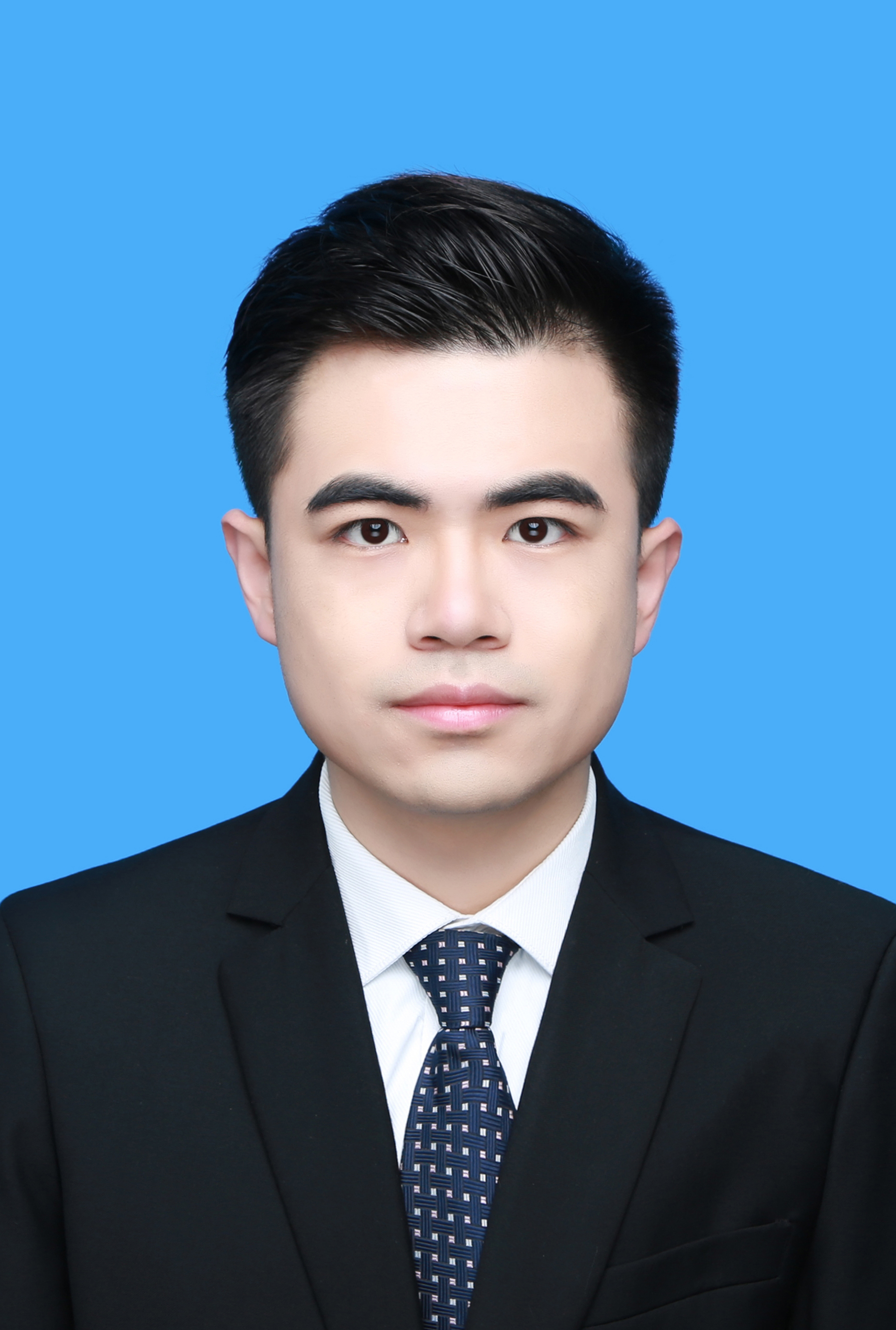}}]
{Jingtao Sun} received the BE and MSc degrees in College of Electrical and Information Engineering from Hunan University, Changsha, China. He is currently working toward the PhD degree with the college of Electrical and Information Engineering from Hunan University.
 His research interests include 3D computer vision, robotics and multi-modal.
\end{IEEEbiography}
\vspace{-0.4cm}
\begin{IEEEbiography}[{\includegraphics[width=1in,height=1.25in,clip,keepaspectratio]{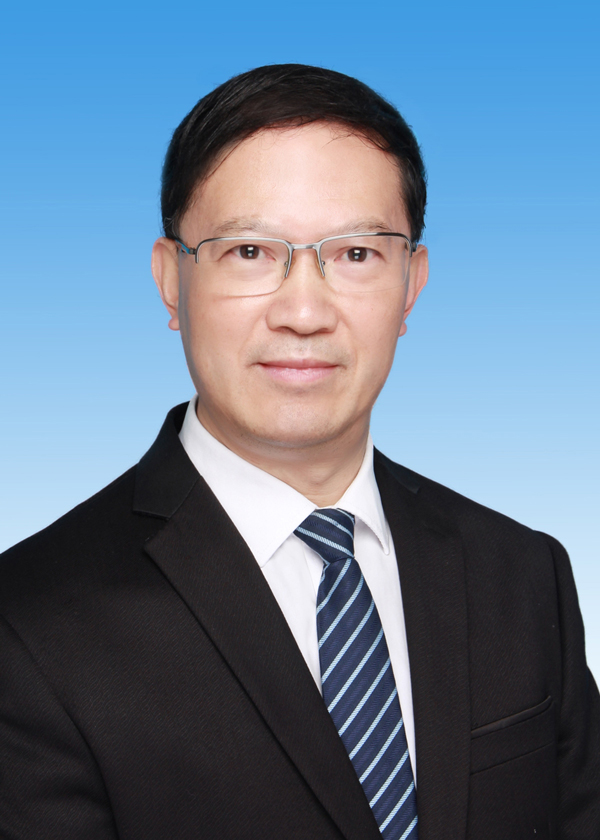}}]
{Yaonan Wang} received the BE degree in computer engineering from East China University of Science and Technology, Fuzhou, China, in 1981 and the MSc and PhD degrees in control engineering from Hunan University, Changsha, China, in 1990 and 1994, respectively. He was a Postdoctoral Research Fellow with the National University of Defense Technology, Changsha, from 1994 to 1995, a Senior Humboldt Fellow in Germany from 1998 to 2000, and a Visiting Professor with the University of Bremen, Bremen, Germany, from 2001 to 2004. He has been a Professor with Hunan University since 1995. His research interests include robotics and computer vision. 
\vspace{-0.4cm}
\end{IEEEbiography}
\begin{IEEEbiography}[{\includegraphics[width=1in,height=1.25in,clip,keepaspectratio]{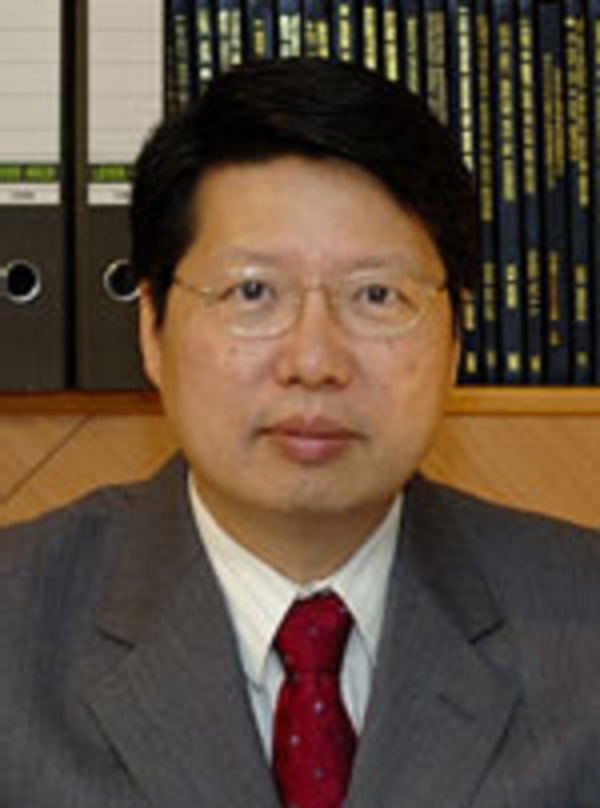}}]
{Danwei Wang} (Life Fellow, IEEE) received the BE degree from the South China University of Technology, China, in 1982, and the MSc and PhD degrees from the University of Michigan, Ann Arbor, MI, USA, in 1984 and 1989, respectively.
He was the Head of the Division of Control and Instrumentation, Nanyang Technological University
(NTU), Singapore, from 2005 to 2011, the Director of the Centre for System Intelligence and Efficiency, NTU, from 2013 to 2016, and the Director of the ST Engineering-NTU Corporate Laboratory, NTU, from 2015 to 2021. He is currently a Professor with the School of Electrical and Electronic Engineering, NTU. His research interests include robotics, robotics vision, and applications. 

\end{IEEEbiography}
\end{document}